\documentclass{article}

\usepackage[subtle]{savetrees}
\usepackage{lipsum,multicol}

\usepackage[utf8]{inputenc} % allow utf-8 input
\usepackage[T1]{fontenc}    % use 8-bit T1 fonts
\usepackage{hyperref}       % hyperlinks
\usepackage{url}            % simple URL typesetting
\usepackage{booktabs}       % professional-quality tables
\usepackage{amsfonts}       % blackboard math symbols
\usepackage{nicefrac}       % compact symbols for 1/2, etc.
\usepackage{microtype}      % microtypography
\usepackage{rotating}

\usepackage[numbers]{natbib}

\usepackage{amsmath}
\usepackage{amsthm}
\usepackage{amssymb}
\usepackage{bm}
\usepackage{mathtools}

\newcommand{\etal}{\textit{et al.}}

\newcommand{\bx}{\boldsymbol{x}}

\newcommand{\boldH}{\boldsymbol{H}}
\newcommand{\boldF}{\boldsymbol{F}}
\newcommand{\bz}{\boldsymbol{z}}

\newcommand{\boldtheta}{{\boldsymbol{\theta}}}

\newcommand{\given}{\,|\,}
\newcommand{\DF}{_{+\mathcal{F}}}
\newcommand{\DmC}{_{-\mathcal{C}}}

\newcommand{\red}[1]{\textcolor{black}{#1}}
\newcommand{\blue}[1]{\textcolor{black}{#1}}

\usepackage{url}

\usepackage[final]{neurips_2022}

\title{Repairing Neural Networks by\\ Leaving the Right Past Behind}
\author{%
  \textbf{Ryutaro Tanno}\thanks{Now at DeepMind, UK. Correspondence to \texttt{rtanno@google.com} and \texttt{yingzhen.li@imperial.ac.uk} }$\ ^{1}$
  \quad \textbf{Melanie F. Pradier}$^1$
  \quad \textbf{Aditya Nori}$^1$
  \quad  \textbf{Yingzhen Li}$^2$ \\
  $^1$Microsoft Health Futures, Cambridge, UK
  \quad\quad $^2$Imperial College London, UK\\
}

\begin{document}
\vspace{-8mm}
\maketitle

\begin{abstract}
%\vspace{-4mm}
Prediction failures of machine learning models often arise from deficiencies in training data, such as incorrect labels, outliers, and selection biases. 
However, such data points that are responsible for a given failure mode are generally not known a priori, let alone a mechanism for repairing the failure. %while maintaining performance on other test examples.
%However, we commonly do not know a priori which datapoints are responsible for a given failure mode. Even if such detrimental training data could be identified, one still needs a mechanism for repairing the failure in a generalisable manner while maintaining performance on other datapoints. 
%
This work draws on the Bayesian view of continual learning, and develops a generic framework for both, identifying training examples which have given rise to the target failure, and fixing the model through erasing information about them. This framework naturally allows leveraging recent advances in continual learning to this new problem of model repairment, while subsuming the existing works on influence functions and data deletion as specific instances. Experimentally, the proposed approach outperforms the baselines for both identification of detrimental training data and fixing model failures in a generalisable manner.
% MEL: Re this last point, I am not sure we beat other baselines in model repairment, should we soften this last sentence in abstract?

\end{abstract}

%%%%%%%%%%%%%%%%%%%%%%%%%%%%%%%%%%%%%%%%%%%%%%%%%%%%%%%%%%%%%%%%%%%%%%%%%
%\vspace{-6mm}
\section{Introduction}
%\vspace{-3mm}
Machine learning (ML) models often exhibit unexpected failures once deployed in the “wild”. Recent lines of research aim to  alleviate different well-known shortcomings in supervised models, such as vulnerability to annotation errors  \citep{frenay2013classification} and adversarial attacks \cite{xu2020adversarial}, sensitivity to data shifts \citep{koh2021wilds}, and biases to underrepresented subgroups \citep{torralba2011unbiased,chen2018my,sagawa2019distributionally}. However, it is often challenging to anticipate beforehand all plausible failure scenarios, and protect against them pre-emptively. This motivates developing a technique that is able to \textit{repair a model on demand}, as new failure cases arise in practice.

Undesirable behaviours of ML models commonly stem from defects in the training data. However, it is unclear how to detect the causes of such failures automatically, rendering a manual troubleshooting necessary. Furthermore, once the problems are uncovered, one would still need to design fixes, which typically involve further data curation/collection, and model retraining/redesigning from scratch. Executing the above steps demand not only time, but also mature expertise in the relevant ML areas, a scarcity in the present job market. 

This work introduces an approach to identifying a set of most detrimental training examples that have caused failure cases observed at test time, and to subsequently repairing the model on these failures by deleting those culprits. At the basis of both cause identification and repairment steps is the approximation of \emph{``counterfactual'' posterior distribution} where some training examples are assumed absent. We formalise this as a Bayesian \emph{continual (un)learning} problem \citep{nguyen2017variational}, where the above counterfactual posterior is estimated by deleting the evidence of selected training data from the current posterior. We note that, while other factors (e.g. model class and optimisation) may play a role, we focus on ``data debugging'' and investigate, to what extent, prediction failures could be remedied by only intervening on data and updating the model accordingly.

\begin{figure}[ht]
\centering
  \vspace{-5mm}
  \includegraphics[width=0.99\linewidth]{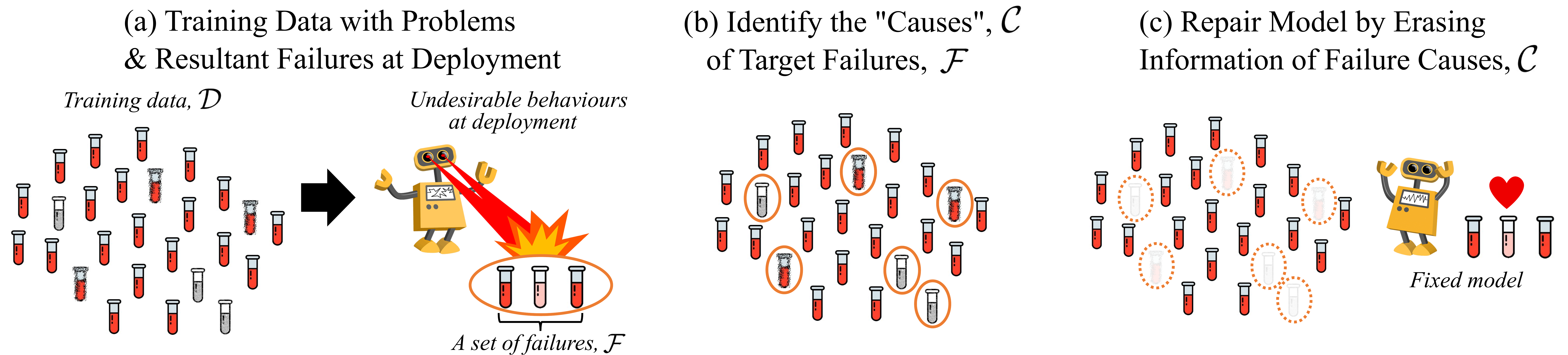}
  \vspace{-3mm}
  \caption{\footnotesize (a) Real-world datasets are often fraught with issues such as annotation noise, low-quality inputs, anomalies, and acquisition biases (e.g. demographic imbalances). Such issues may lead to undesirable performance of the trained models in deployment. Our approach aims at repairing such models by (b) identifying detrimental training examples which have caused the target failures, and then (c) erasing efficiently the \textit{memories} of those examples from the models.}
  \label{fig:overview}
  \vspace{-2mm}
\end{figure}

%\RT{Also explain the proposed two-step approach as depicted in Fig.~\ref{fig:overview}}.
Fig.~\ref{fig:overview} gives an overview of the proposed approach for model repairment, which operates in two steps by 1) identifying causes of failure among training data, and 2) updating the model by erasing the ``memories'' of those harmful examples. Importantly, the proposed framework is agnostic to \emph{why} a particular datapoint hurts model performance, handling both, issues in the input data and/or labels. We only require specification of the set of failed cases for which we wish to improve performance.

% TODO: make contributions more concise, come back once experiments have been written
\textbf{Our contributions:} 
We develop a framework for \textit{repairing} machine learning models by erasing memories of detrimental datapoints. The framework connects both identification and removal of detrimental data under a (Bayesian) continual learning perspective, which brings forth practical benefits. Firstly, the framework subsumes works on influence function \cite{koh2017understanding} and data deletion \cite{guo2019certified} as specific examples, which are developed independently, and our work reveals their close connections and limitations. Secondly, the generality of our formulation allows translating any continual learning method into this model repairment setting, and opens doors to further research. In particular, we extend Elastic Weight Consolidation \citep{kirkpatrick2017overcoming} -- a specific continual learning algorithm -- to cause identification and data removal, and demonstrate improvements over the prior works in a variety of settings where training data are contaminated with annotation and/or input noise. 
% This work formalises the task of \textit{model repairment}, and develops a generic framework for this new task. The framework connects both identification and removal of detrimental data under a (Bayesian) continual learning perspective, which enables leveraging advances in continual learning for model repairment. This is demonstrated by: 1) a novel development that extends Elastic Weight Consolidation \citep{kirkpatrick2017overcoming} -- a specific continual learning algorithm -- to cause identification and information removal; and 2) a formal discussion on the linear influence function \citep{koh2017understanding} as another specific instance, which shows the generality of our framework. Empirical studies show the utility of the proposed approach for model repairment in a variety of settings in which training data are contaminated with annotation and/or input noise. 
\vspace{-1mm}

\section{Model Repairment by Data Deletion} \label{sec:method}
%\vspace{-3mm}
Let us consider a prediction model $p(y \given \bx, \boldtheta)$ that returns a probability distribution of the output $y$ given an input $\bx$.
% By denoting $p(\bz \given \boldtheta) = p(y \given \boldtheta, \bx)$ for $\bz = (\bx, y)$, and making the i.i.d.~assumption, after observing training data $\mathcal{D}=\{\bz^{(n)} = (\bx^{(n)}, y^{(n)})\}_{n=1}^N$, the posterior distribution over its parameters is $p(\boldtheta \given \mathcal{D}) = \frac{\prod_{\bz \in \mathcal{D}} p(\bz \given \boldtheta) p(\boldtheta)}{p(\mathcal{D})}$.
We make the i.i.d.~modelling assumption and denote $p(\bm{\theta} \given \mathcal{D})$ as the posterior distribution over the model parameters $\bm{\theta}$ given training data $\mathcal{D}=\{(\bx^{(n)}, y^{(n)})\}_{n=1}^N$.
%Here $p(\boldtheta)$ is the prior distribution on $\boldtheta$ and $p(\mathcal{D}) = \int \prod_{\bz \in \mathcal{D}} p(\bz \given \boldtheta) p(\boldtheta) d \boldtheta$ is the marginal likelihood of our dataset $\mathcal{D}$.
%
At test time, the posterior predictive distribution is used to infer label $y^\star$ given a new sample $\bx^\star$:
\begin{equation}\label{eq:predictive_inference}
    p(y^\star|\bm{x}^\star, \mathcal{D}) = \int p(y^\star \given \bm{x}^\star, \bm{\theta}) p(\bm{\theta} \given \mathcal{D}) d\bm{\theta},
\end{equation}
where $p(y^\star |\bm{x}^\star, \bm{\theta})$ is the likelihood term for sample $(x^\star, y^\star)$.
While we use approximate posteriors in practice, we focus on the exact case for now to formalise the problem.

Imagine that, in deployment, this model makes incorrect predictions in certain situations. After collecting a ``failure set'' $\mathcal{F} = \{(\bx_{f}^{(n)}, y_{f}^{(n)})\}_{n=1}^{N_f}$ of examples from such failure mode in the test set\footnote{There may be multiple different ways in which the model fails \cite{henn2021principled}, and the failure cases may consist of several groups. One type of mistake might incur more costs than others (e.g. in certain medical applications, \textit{false negative} is more costly than \textit{false positive}). Here we assume that we have identified at least one failure type that we would like to fix.}, we would like to \textit{repair} the model, such that it improves performance on the failure set $\mathcal{F}$ and similar future cases. We also argue that a successful repairment should also \textit{maintain} a similar level of performance on the rest of test examples.
These two objectives of model repairment are analogous to those of a \textit{medical treatment}, in that both aim to fix a specific problem while leaving the ``healthy'' part as intact as possible. A further discussion of different aspects, including generalisation, efficiency and specificity properties is provided in Appendix \ref{app:problem_formulation}.

In this work, we assume that the main reason for such failures $\mathcal{F}$ is due to the existence of detrimental examples in the training data $\mathcal{D}$, for example, noisy labels, low-quality inputs and/or group imbalances. Our hypothesis is that, by removing these harmful datapoints and adapting the model accordingly, the model can be repaired to return correct predictions for datapoints in $\mathcal{F}$ as well as similar future cases. We acknowledge that there might be other reasons for a model to make wrong predictions on $\mathcal{F}$, such as bad local optima and model biases, which are outside the scope of this work. Addressing data-based failures is complementary to accounting for other problems and is of relevance regardless, as datasets typically come with unexpected issues no matter how much we curate them beforehand.
% MEL: Think of a sentence to defend this a little bit more, those other works and this one complement each other, and datasets will always come with issues, so we need our approach - DONE
% MEL: last sentence I wrote is a little wrong, consider moving to Discussion/conclusion?

We formulate the process of \textbf{model repairment} as the following two steps (see Fig.~\ref{fig:overview} for an illustration): 
\begin{enumerate}
    \item \textbf{Cause identification:} Identify a set of detrimental datapoints, i.e.~``failure causes'' $\mathcal{C}$ in the training data $\mathcal{D}$ that contributed the most to the failure set $\mathcal{F}$.

    \item \textbf{Treatment: } Given the set of failure causes $\mathcal{C}$, adapt the model to predict correctly on the failure set $\mathcal{F}$, while maintaining performance on remaining test examples. 
\end{enumerate}
Below we describe a formal framework for performing these two steps. In the first phase of cause identification, we need to define a measure of how much a subset of training examples $\mathcal{C} = \{(\bx_{c}^{(n)}, y_{c}^{(n)})\}_{n=1}^{N_c}\subset \mathcal{D}$ is responsible for the failure cases $\mathcal{F}$. To this end, we propose to check how much the posterior predictive distribution on $\mathcal{F}$ changes as a result of deleting $\mathcal{C}$ from the training data: 
\vspace{0mm}
\begin{equation}
r(\mathcal{C}) \coloneqq \log \, p(\mathcal{F} \given \mathcal{D}\setminus \mathcal{C}) - \log \, p(\mathcal{F} \given \mathcal{D}), \label{Eq.rC}
\end{equation}
where $p(\mathcal{F} \given \mathcal{D})$ and $p(\mathcal{F} \given \mathcal{D}\setminus \mathcal{C})$ are the \blue{posterior predictive distributions} before and after removing a subset of training examples $\mathcal{C}$ defined as follows:
\begin{equation}
%\hspace{-1mm}
    p(\mathcal{F} \given \mathcal{D}) = \int p(\mathcal{F} \given \boldtheta) p(\boldtheta|\mathcal{D}) d{\boldtheta}, \ \ p(\mathcal{F} \given \mathcal{D}\setminus \mathcal{C}) = \int p(\mathcal{F} \given \boldtheta) p(\boldtheta|\mathcal{D}\setminus \mathcal{C}) d{\boldtheta}, \ \
    p(\mathcal{F} \given \boldtheta) = \prod_{(\bm{x}, y) \in \mathcal{F}} p(y \given \bm{x}, \boldtheta).
\end{equation}
%
%\footnote{Assuming that test datapoints are independent, note that the denominator factorises over the individual failure cases: $$p(\mathcal{F}| \mathcal{D}) = \prod_{(\bx^*, y^*) \in \mathcal{F}}p(y^*|\bm{x}^*, \mathcal{D}),$$ where each predictive posterior distribution on the RHS is obtained as described in eq.~\eqref{eq:predictive_inference}. The numerator distribution $p(\mathcal{F}| \mathcal{D}\setminus \mathcal{C})$ after removal of $\mathcal{C}$, also factorises similarly.\YL{YL: this claim of factorisation is problematic, I suggest removing this footnote.}
If a given $\mathcal{C}$ leads to a large positive value of $r(\mathcal{C})$ in Eq.~\eqref{Eq.rC}, it means that removing $\mathcal{C}$ from $\mathcal{D}$ would have improved the performance of the Bayesian predictive inference on the failure set $\mathcal{F}$. The first \textit{cause identification} step thus entails finding a subset $\mathcal{C}$ with the maximal log-density ratio $r(\mathcal{C})$. In the second \textit{treatment} step, we can directly adopt $p(\mathcal{F} \given \mathcal{D}\setminus \mathcal{C})$ for such identified $\mathcal{C}$ as the updated predictive distribution, as it confers the largest improvement over the failure cases. In other words, the posterior predictive distribution $p(\mathcal{F} \given  \mathcal{D}\setminus \mathcal{C})$ after data removal is key to both the search of detrimental datapoints and repairment of the model. 

The central computational question is, therefore, concerned with the \textit{efficient calculation of $p(\mathcal{F} \given \mathcal{D}\setminus \mathcal{C})$ without retraining the model from scratch}. The reasons for avoiding retraining are: (1) in the cause identification step, it is computationally prohibitive as retraining needs to be done for every subset $\mathcal{C} \subset \mathcal{D}$; and (2) in the treatment step, the resulting retrained model may be drastically different from the original model one would like to fix (due to, e.g.~noises in the SGD-based optimisation and parameter re-initialisation), and may lose other desirable properties that one may wish to retain.

\red{In this work, we present a continual learning \cite{parisi2019continual} framework to simultaneously address the aforementioned computational challenges in cause identification and treatment. The framework is summarised in Algorithm~\ref{alg:cause_identification}, with the following key developments by leveraging (Bayesian) continual learning:}
\begin{enumerate}
    \item For \textbf{cause identification} (Section \ref{sec:cause_identification}), we present a fast approximation to $r(\mathcal{C})$, which requires a one-off approximation of $p(\boldtheta \vert \mathcal{D}, \mathcal{F})$ only via continual learning, and enables a \emph{linear-time} search for the detrimental datapoints in $\mathcal{C}$.

    \item For \textbf{treatment} (Section \ref{sec:treatment}), we show that the approximation of $p(\boldtheta \vert \mathcal{D}\setminus \mathcal{C})$ for $\mathcal{C}$ identified in the first step can be achieved by performing a new ``continual learning task'' using $\log p(\mathcal{C} \given \boldtheta)$ as its ``loss function''.
\end{enumerate}
Our framework is generic and flexible in the sense that any continual learning approach can be applied to both steps; we demonstrate, in particular, a concrete instantiation by leveraging Elastic Weight Consolidation \citep{kirkpatrick2017overcoming} as the base continual learning approach. We now elaborate on the mathematical details of the two steps.

\begin{algorithm}[t]
  
   \caption{\footnotesize Model Repairment}
   \label{alg:cause_identification}
\begin{algorithmic}
    \footnotesize
   \STATE {\bfseries Input:} training data $\mathcal{D}$; failure cases $\mathcal{F}$; approximate posterior $q(\boldtheta) \approx  p(\boldtheta \given \mathcal{D})$; likelihood $p(\bz \given \boldtheta)$
   \STATE {\bfseries Output:} failure causes $\mathcal{C}$, ``repaired'' posterior $q\DmC(\boldtheta)$
%   \STATE
   \STATE {\bfseries \footnotesize \textcolor{blue}{\texttt{\# Step I: Cause Identification}}}
   \STATE {\bfseries Update posterior:} Apply a \textit{continual learning} method to obtain $q\DF(\boldtheta) \approx p(\boldtheta \vert \mathcal{D}, \mathcal{F})$ by fitting the failure set $\mathcal{F}$
 
   \STATE {\bfseries Compute influences of training examples on $\mathcal{F}$:} Calculate $\tilde{r}(\bz)\,\, \forall \bz \in \mathcal{D}$ (Eq.~\eqref{eq:final_influence_score})
   
   \STATE {\bfseries Find failure causes $\mathcal{C}$:} Return the examples with positive influence, $\mathcal{C} \leftarrow \{\bz \in \mathcal{D}$: $\tilde{r}(\bz) > 0\}$ 
%   Initialise $\mathcal{C} = \emptyset$, then
%   \FOR{$\bz \in \mathcal{D}$}
%   \IF{$\tilde{r}(\bz) > 0$}
%   \STATE Update $\mathcal{C} \leftarrow \mathcal{C}\cup \{\bz\}$
%   \ENDIF
%   \ENDFOR
%   \STATE 
   \STATE {\bfseries \footnotesize \textcolor{red}{\texttt{\# Step II: Treatment}}}
   \STATE {\bfseries Delete information of $\mathcal{C}$}: Apply a \textit{continual (un)learning} method to the original posterior $q(\boldtheta)$, and obtain the posterior on the corrected data $q\DmC(\boldtheta) \approx p(\boldtheta \vert \mathcal{D}\setminus \mathcal{C})$
\end{algorithmic}
\vspace{-1mm}
\end{algorithm}

%\vspace{-5mm}
\subsection{Step I: Cause Identification}\label{sec:cause_identification}
%\vspace{-2mm}
%This problem introduces multiple approximation challenges. \RT{1) Recomputing the posterior $P(\mathcal{F}| \mathcal{D}\setminus \mathcal{C}) $ for every possible $\mathcal{C}$; 2).Combinatorial search problem, 3)Approximation of the posterior }. 

% \YL{The derivation of eq (4) relies on i.i.d.~assumptions, i.e.~at this stage we already need to assume $p(\mathcal{D}, \boldtheta) = \prod_{\bm{z} \in \mathcal{D}} p(\bm{z} | \boldtheta) p(\boldtheta)$. I suggest we make this clear at the very beginning of section 2, before eq. (1).}

Identifying the set of detrimental examples $\mathcal{C}$
% MEL: you are using C to design any arbitrary set previously, and now to refer to the set of harmful datapoints. So maybe replace by C^\dagger for the harmful set, and keep C for any arbitrary set instead of C'?
requires solving the following optimisation problem
\begin{equation}
\mathcal{C} = \operatorname{argmax}_{\mathcal{C}' \in \mathbb{P}(\mathcal{D})}  r(\mathcal{C}'),
\end{equation}
where $\mathbb{P}(\mathcal{D})$ denotes the power set of $\mathcal{D}$. Solving this comes with multiple computational challenges. %Algorithm~\ref{alg:cause_identification} summarises the whole procedure.
% MEL: consider moving Algorithm reference after explanation of the Algorithm.

Firstly, a naive approach would require computing the predictive distribution $p(\mathcal{F} \given \mathcal{D}\setminus \mathcal{C})$ --- and thus, the posterior $p(\boldtheta \given \mathcal{D} \setminus \mathcal{C})$ --- for every subset $\mathcal{C}$ of $\mathcal{D}$, which is prohibitively expensive. To address this, we present a \textbf{``predictive'' approach} that removes this computational burden. 
The key idea is to notice that 
$$p(\mathcal{D} \setminus \mathcal{C} \given \boldtheta) = p(\mathcal{D} \given \boldtheta) / p(\mathcal{C} \given \boldtheta), \quad \forall \mathcal{C} \subset \mathcal{D},$$ 
due to the i.i.d.~modelling assumption. Inserting this into the Bayes' rule for computing $p(\boldtheta|\mathcal{D} \setminus \mathcal{C})$ yields:
\begin{equation*}
\begin{aligned}
\log p(\mathcal{F} \given \mathcal{D} \setminus \mathcal{C}) &= \log \int p(\mathcal{F} \given \boldtheta) p(\boldtheta \given \mathcal{D} \setminus \mathcal{C}) d\boldtheta 
%&= \log \int \frac{p(\mathcal{D})}{p(\mathcal{D}\setminus \mathcal{C})}\cdot \frac{p(\mathcal{F} \given \boldtheta) p( \mathcal{D}, \boldtheta)}{p(\mathcal{D})p(\mathcal{C} \given \boldtheta)} d\boldtheta \\
= \log \int \frac{p(\mathcal{F} \given \boldtheta)}{p(\mathcal{C}\vert \boldtheta)} \frac{p(\mathcal{D} |\boldtheta) p(\boldtheta)}{p(\mathcal{D})} d\boldtheta - \log\ \frac{p(\mathcal{D}\setminus \mathcal{C})}{p(\mathcal{D})}.
\end{aligned}
\end{equation*}
% MEL: Add Eq. number, decrease Equation font
Notice that again due to the i.i.d.~modelling assumption,
$$ \frac{p(\mathcal{D}\setminus \mathcal{C})}{ p(\mathcal{D})} = \int \frac{1}{p(\mathcal{C} \given \boldtheta)} \frac{p(\mathcal{D} \given \boldtheta) p(\boldtheta)}{p(\mathcal{D})} d\boldtheta.$$ 
Therefore, we can compute the log density ratio $r(\mathcal{C}) = \log \,  p(\mathcal{F} \given \mathcal{D}\setminus \mathcal{C}) - \log \, p(\mathcal{F} \given \mathcal{D})$ in the following ``predictive'' form, without computing $p(\boldtheta \given \mathcal{D} \setminus \mathcal{C})$---see Appendix \ref{sec:supp_derivations} for derivations:
\begin{equation} \label{eq:predictive_log_density_ratio}
r(\mathcal{C})= \log \int \frac{ p(\boldtheta \given \mathcal{D}, \mathcal{F})}{p(\mathcal{C} \given \boldtheta)} d\boldtheta - \log \int \frac{p(\boldtheta \given \mathcal{D})}{p(\mathcal{C} \given \boldtheta)}  d\boldtheta = \log\ \mathbb{E}_{p(\boldtheta \given \mathcal{D}, \mathcal{F})}[p(\mathcal{C} \given \boldtheta)^{-1}] - \log\ \mathbb{E}_{p(\boldtheta \given \mathcal{D})}[p(\mathcal{C} \given \boldtheta)^{-1}]. 
\end{equation}
\red{In this form, we only need to compute the posterior $p(\boldtheta| \mathcal{D}, \mathcal{F})$ once, and can side-step the requirement of computing $p(\boldtheta|\mathcal{D} \setminus \mathcal{C})$ for inspection of every new $\mathcal{C} \subset \mathcal{D}$, thereby removing one of the key computational bottlenecks. 
Intuitively, as the ``predictive'' formulation (Eq.~\eqref{eq:predictive_log_density_ratio}) of $r(\mathcal{C})$ computes the log expectation difference of $p(\mathcal{C} \given \boldtheta)^{-1}$, higher $r(\mathcal{C})$ means that datapoints in $\mathcal{C}$ are less likely to be predicted correctly when the posterior is updated by the information from the failure set $\mathcal{F}$. In other words, this indicates conflicting information exists between $\mathcal{F}$ and $\mathcal{C}$, and as we would like to repair the model to produce correct predictions on $\mathcal{F}$ and similar examples, the information of $\mathcal{C}$ should be removed from the model.}

Secondly, we are still left with the combinatorial search for the best subset $\mathcal{C}$, which is also prohibitive when the size of training data $\mathcal{D}$ is large, even when the ``predictive'' formulation Eq.~\eqref{eq:predictive_log_density_ratio} is used. \blue{We address this issue by a first-order Taylor series approximation of $r(\mathcal{C})$. Let us re-write the log density ratio as
%%%%%%%%%%%%%%%%%%%%%%%%%%%%%%%%%%%%
%% Jensen's inequality (wrong!) 
%%%%%%%%%%%%%%%%%%%%%%%%%%%%%%%%%%%%
% Specifically, by using Jensen's inequality, we have that:
% \begin{equation}\label{eq:log_density_approximation}
%     \begin{aligned}
%     r(\mathcal{C}) &\geq \int \log \frac{ p(\boldtheta \vert \mathcal{D}, \mathcal{F})}{p(\mathcal{C}\vert \boldtheta)} d\boldtheta - \int \log \frac{p(\boldtheta \vert \mathcal{D})}{p(\mathcal{C}\vert \boldtheta)}  d\boldtheta \\ 
%     &= \mathbb{E}_{p(\boldtheta | \mathcal{D})}\left[\log p(\mathcal{C}\vert \boldtheta) \right] - \mathbb{E}_{p(\boldtheta | \mathcal{D}, \mathcal{F})}\left[\log p(\mathcal{C}\vert \boldtheta) \right] \\ &=: \hat{r}(\mathcal{C}). 
%     \end{aligned}
% \end{equation}
%%%%%%%%%%%%%%%%%%%%%%%%%%%%%%%%%%%%%%%%%%%%%%%%%%%%%%%%%%%%%%%%
%%% Another way of deriving the approximation (based on a Taylor approximation) %%%%
%%%%%%%%%%%%%%%%%%%%%%%%%%%%%%%%%%%%%%%%%%%%%%%%%%%%%%%%%%%%%%%%%%%%%%%%%%%%%
\begin{align}
    r(\mathcal{C}) = F\left(1, \, p\left(\boldtheta \given \mathcal{D}, \mathcal{F}\right)\right) - F\left(1,  \,p\left(\boldtheta \given \mathcal{D}\right)\right),\label{Eq:rCF} \quad \text{where}\quad 
    F(\epsilon, \,g(\boldtheta)) \coloneqq   \log \int  g(\boldtheta) e^{-\epsilon \log p(\mathcal{C} \given \boldtheta)} d\boldtheta.%\label{Eq:Ffunction}
\end{align}
Note that $F(0, g(\boldtheta)) = 0$ for any well-defined distribution $g(\boldtheta)$. We now perform a Taylor expansion of $F(\epsilon, g(\boldtheta))$ around $\epsilon = 0$:
$F(\epsilon, g(\boldtheta)) = - \epsilon \, \mathbb{E}_{ g(\boldtheta)} \left[\log p(\mathcal{C} \given \boldtheta) \right] + \mathcal{O}(\epsilon^2)$.
Finally we obtain an approximate log-density ratio $\hat{r}(\mathcal{C}) \approx r(\mathcal{C})$ by plugging the first term of the Taylor expansion into the RHS of Eq.~(\ref{Eq:rCF})}:
\begin{equation}\label{eq:log_density_approximation}
\hat{r}(\mathcal{C}) \coloneqq \mathbb{E}_{p(\boldtheta \given \mathcal{D})}\left[\log p(\mathcal{C}\given \boldtheta) \right] - \mathbb{E}_{p(\boldtheta \given \mathcal{D}, \mathcal{F})}\left[\log p(\mathcal{C}\given \boldtheta) \right].
\end{equation}
%
%
%%%%%%%%%%%%%%%%% End of the alternative derivation %%%%%%%%%%%%%%%%%%% 
%
Assuming that data are i.i.d., and defining $\bz = (\bx, y)$, the above approximation can be expressed as the sum of individual log density ratios, $\hat{r}(\mathcal{C}) =  \sum_{\bz\in\mathcal{C}}\hat{r}(\bz)$, where each term is given by
\begin{equation}
    \begin{aligned}
    \hat{r}(\bz)= \mathbb{E}_{p(\boldtheta \given \mathcal{D})}\left[\log p(\bz \given \boldtheta) \right] - \mathbb{E}_{p(\boldtheta \given \mathcal{D}, \mathcal{F})}\left[\log p(\bz \given \boldtheta) \right], \quad p(\bz \given \boldtheta) = p(y \given \bx, \boldtheta).
    \end{aligned}  
\label{eq:predictive_difference_per_instance}
\end{equation}
%where $p(\bz \given \boldtheta) = p(y \given \bx, \boldtheta)$.
%
Critically, with this approximation, in order to find a subset $\mathcal{C}$ of cardinality $K$ that leads to the maximal $\hat{r}(\mathcal{C})$, it suffices to compute $\hat{r}(\bz)$ for every training example $\bz \in \mathcal{D}$ and find the top $K$ examples with largest $\hat{r}(\bz)$ values, thereby reducing the search space from $\mathcal{O}(|\mathcal{D}|!)$ choices to only $\mathcal{O}(|\mathcal{D}|)$ choices.

The per-instance formulation Eq.~\eqref{eq:predictive_difference_per_instance} has similar interpretation as the ``predictive'' formula Eq.~\eqref{eq:predictive_log_density_ratio}, in that $\hat{r}(\bz)$ measures how much the predictive moments of $\bz$ changes when the model is further trained on $\mathcal{F}$, i.e., computation of $p(\boldtheta \given \mathcal{D}, \mathcal{F})$. If the difference is positive, i.e., $\hat{r}(\bz) > 0$, it means the example $\bz \in \mathcal{D}$ is a conflicting evidence against the test examples in $\mathcal{F}$; conversely, if $\hat{r}(\bz) < 0$, then $\bz$ and $\mathcal{F}$ are aligned. In practice, the failure causes correspond to examples $\bz$ with $\hat{r}(\bz) > 0$. 

%See Algorithm~\ref{alg:cause_identification} for details.

Lastly, for non-linear models (e.g.~neural networks), approximate posteriors $q(\boldtheta) \approx  p(\boldtheta \given \mathcal{D})$ and $q\DF(\boldtheta) \approx p(\boldtheta \given \mathcal{D}, \mathcal{F})$ are needed due to intractability of the exact posteriors. We assume that $q(\boldtheta)$ is available after training, and suffers from the prediction failures $\mathcal{F}$. As recomputing the posterior $q\DF(\boldtheta)$ from scratch can be expensive, we propose to use a \textit{continual learning} technique \cite{parisi2019continual} and obtain this quantity by updating the original posterior $q(\boldtheta)$. Finally, we use the following metric  $\tilde{r}(\bz)$ in practice to calculate the detrimental impact of each training datapoint on the failure set, $\mathcal{F}$, by replacing the exact posteriors in Eq.~\eqref{eq:predictive_difference_per_instance} with their corresponding approximations: 
\begin{equation}\label{eq:final_influence_score}
    \tilde{r}(\bz):= \mathbb{E}_{q(\boldtheta)}\left[\log \, p(\bz\vert \boldtheta) \right] - \mathbb{E}_{q\DF(\boldtheta) }\left[\log \, p(\bz\vert \boldtheta) \right],
\end{equation}
% \vspace{-5mm}
and the top $K$ entries according to $\tilde{r}(\bz)$ are selected to approximate the failure causes $\mathcal{C}$. This metric is generic, and its implementation depends on the specifics in which both $q(\boldtheta)$ and $q\DF(\boldtheta)$ are computed, e.g., MLE/MAP point estimates, Laplace approximation \citep{mackay1992bayesian}, variational inference \citep{jordan1998introduction,blundell2015weight}, etc. 
Two concrete examples are provided: the first shows that the well-known \textit{linear influence function} \cite{koh2017understanding} is a specific instance of Eq.~\eqref{eq:final_influence_score}; and the second is derived by extending a continual learning method, known as \textit{Elastic Weight Consolidation} (EWC) \cite{kirkpatrick2017overcoming} to cause identification, and is a key methodological development in our work. 

\textbf{\textit{Example 1 (Linear Influence Function):}}
%In their seminal work, \citet{koh2017understanding}, introduced \textit{linear influence functions} to measure the effects of training samples on test-time predictions of non-linear models. 
Our proposed metric in Eq.~\eqref{eq:final_influence_score} recovers the \emph{linear influence function} from Koh \& Liang \citep{koh2017understanding} when point estimates are used for $\boldtheta$. Assume that the model is trained on data $\mathcal{D}$ with parameters $\hat{\boldtheta}$, which corresponds to an approximation of MLE/MAP estimates, i.e., $q(\boldtheta) = \delta(\boldtheta - \hat{\boldtheta}) \approx p(\boldtheta \given \mathcal{D})$. After observing the set of failures $\mathcal{F}$, a point estimate of $p(\boldtheta \given \mathcal{D}, \mathcal{F})$ is obtained by performing a single update of natural gradient ascent \citep{amari1998natural} on the log likelihood of $\mathcal{F}$ with step size $\gamma>0$:
\begin{equation}\label{eq:linear_influence}
q\DF(\boldtheta) = \delta(\boldtheta - \hat{\boldtheta}\DF) \approx p(\boldtheta \given \mathcal{D}, \mathcal{F}), \quad
\hat{\boldtheta}\DF \approx \hat{\boldtheta} + \gamma \hat{\boldF}_{\hat{\boldtheta}}^{-1}\nabla_{\hat{\boldtheta}}\log p(\mathcal{F} \vert \hat{\boldtheta}),
\end{equation}
where $\hat{\boldF}_{\hat{\boldtheta}}$ is the empirical Fisher information matrix. This means 
\begin{equation} \label{eq:predictive_point_estimates}
 \tilde{r}(\bz) = - \gamma \nabla_{\hat{\boldtheta}}\log p(\mathcal{F}\vert\hat{\boldtheta})^{\top} \hat{\boldF}_{\hat{\boldtheta}}^{-1}\nabla_{\hat{\boldtheta}}\log p(\bz\vert\hat{\boldtheta}),
\end{equation}
if defining the updated posterior as $q\DF(\boldtheta) = \delta(\boldtheta - \hat{\boldtheta}\DF)$.
The negation of the above equation coincides with the linear influence function (Eq.~(2) in \cite{koh2017understanding}) when the failure set is assumed to be \blue{a singleton $\mathcal{F} = \{\bz_{test}\}$ and $\gamma=1$. Note that the sign difference arises since our work aims to quantify the ``negative'' influence rather than the "positive" one in contrast with the work of Koh and Liang \cite{koh2017understanding}}. 

\textit{\textbf{Example 2 (Elastic Weight Consolidation): }}
The generality of Eq.~\eqref{eq:final_influence_score} permits any continual learning method of one's choice for estimating the updated posterior $q\DF(\boldtheta) \approx p(\boldtheta \given \mathcal{D}, \mathcal{F})$ after observing failure samples $\mathcal{F}$. Here we illustrate how EWC 
\citep{kirkpatrick2017overcoming} as a continual learning method can be adopted in the context of cause identification. EWC approximates $ p(\boldtheta \given \mathcal{D}, \mathcal{F})$  by first performing Laplace approximation of the original posterior $p(\boldtheta \given \mathcal{D})$ around the point estimate $\hat{\boldtheta}$, and subsequently finding the MAP solution of $\boldtheta$. Formally, $\hat{\boldtheta}\DF$ is obtained by maximising the objective below w.r.t. $\boldtheta$ via SGD (see Appendix \ref{app:ewc_identification_objective} for details): 
\begin{equation}\label{eq:predictive_ewc}
\log \, p(\mathcal{F} \vert\boldtheta)  - \frac{N}{2}(\boldtheta - \hat{\boldtheta})^\top  \hat{\boldF}_{\hat{\boldtheta}} (\boldtheta - \hat{\boldtheta}) - \frac{\lambda}{2}||\boldtheta - \hat{\boldtheta}||^2_2,
\end{equation}
where the off-diagonal elements of $\hat{\boldF}_{\hat{\boldtheta}}$ are dropped for memory reason in practice. \blue{Intuitively, the first term encourages high accuracy on the failure set $\mathcal{F}$ whilst the second/third terms ensures the model parameters do not deviate too much in distribution.} Defining $q(\boldtheta) = \delta(\boldtheta - \hat{\boldtheta})$ and $q\DF(\boldtheta) = \delta(\boldtheta - \hat{\boldtheta}\DF)$, we have that
\begin{equation}\label{eq:ewc_influence_function}
 \tilde{r}(\bz) = \log p(\bz \given \hat{\boldtheta}) - \log p(\bz \given \hat{\boldtheta}\DF).
\end{equation}
% MEL: Consider using \tilde{r}_{\text{EWC}}(\bz) and \tilde{r}_{\text{DF}}(\bz)
To compute the above for each datapoint $\bz \in \mathcal{D}$, we only need to solve the optimisation problem of Eq.~(\ref{eq:predictive_ewc}) by SGD once. We refer to this version of $\tilde{r}(\bz)$ as \textit{EWC-influence function}.  
%
% % Commented out for now as it's unlikely to have results on this for this submission. But I'll implement it after ICML. 
% \RT{Alternatively, we can also factor in the variance information in $\tilde{r}(\bz)$ by performing Laplace approximation of $p(\boldtheta \vert \mathcal{D}, \mathcal{F})$ around $\boldtheta^{*}$, and considering $q(\boldtheta) = \mathcal{N}(\boldtheta;\hat{\boldtheta}, \hat{\boldF}_{\hat{\boldtheta}}^{-1})$ and $q\DF(\boldtheta) = N(\boldtheta; \hat{\boldtheta}^{*}, \hat{\boldF}_{\hat{\boldtheta}^{*}}^{-1})$. In this case, we use Monte Carlo sampling to estimate the moment difference in Eq.~\eqref{eq:final_influence_score} by drawing samples from $q(\boldtheta)$ and $q\DF(\boldtheta)$. We refer to this version as \textit{EWC-influence-v2}.} 

% \YL{YL: where does the $n/2$ coefficient in Eq.~(\ref{eq:predictive_ewc}) come from?}\RT{Ah, it comes from the second order of the Taylor approximation I think.}

\textbf{Comparison:}
The EWC-influence function generalises the linear influence approach. To see this, we derive the fixed point of the EWC objective Eq.~(\ref{eq:predictive_ewc}) w.r.t.~$\boldtheta$ (where we set $\lambda = 0$):
\begin{equation}
    \boldtheta = \hat{\boldtheta} + N^{-1}
    \hat{\boldF}_{\hat{\boldtheta}}^{-1} \nabla_{\boldtheta} \log p(\mathcal{F} \vert\boldtheta).
\end{equation}
Then the update in Eq.~(\ref{eq:linear_influence}) that is implicitly used by linear influence function can be viewed as (damped) one-step fixed-point iteration update initialised at $\hat{\boldtheta}$ for solving the fixed-point equation. As EWC-influence update (Eq.~(\ref{eq:ewc_influence_function})) is obtained by using the optimum of Eq.~(\ref{eq:predictive_ewc}), it is arguably more accurate than linear influence function (Eq.~(\ref{eq:predictive_point_estimates})) for measuring the (detrimental) effect of a datum $\bm{z}$ to the model failures $\mathcal{F}$. 

% % \subsubsection*{Probability of Sufficiency \cite{chakarov2016debugging} as a specific instance:}

\subsection{Step II: Treatment} \label{sec:treatment}
%\vspace{-1mm}
Once the causes $\mathcal{C}$ of the failures $\mathcal{F}$ are identified among the training data $\mathcal{D}$, we seek to repair the model $q(\bm{\theta})$ by erasing the memories of $\mathcal{C}$. We formalise this problem as the computation of the posterior $p(\boldtheta \given \mathcal{D}\setminus \mathcal{C})$, i.e., $\mathcal{C}$ is absent from the training data $\mathcal{D}$.
A naive approach would re-run approximate inference on the whole ``corrected'' dataset $\mathcal{D}\setminus \mathcal{C}$ to obtain an approximate posterior $q\DmC(\bm{\theta}) \approx p(\bm{\theta} \given \mathcal{D}\setminus \mathcal{C})$ which can be time consuming. But more fundamentally, by doing so, the obtained $q\DmC(\bm{\theta})$ may be unrelated to the original $q(\bm{\theta})$ based on which $\mathcal{C}$ were identified, due to, e.g.~non-convex SGD optimisation issues. Moreover, this ``model replacement'' approach may not maintain other good properties of the original model $q(\bm{\theta})$.

Analogous to cause identification (Sec.~\ref{sec:cause_identification}), we propose to employ the continual learning approach to estimate efficiently the modified posterior $p(\boldtheta \given \mathcal{D}\setminus \mathcal{C})$. Applying Bayes' rule and some algebraic manipulations yield 
$p(\boldtheta \given \mathcal{D}\setminus \mathcal{C}) \propto p(\boldtheta \given \mathcal{D}) / p(\mathcal{C} \given \boldtheta)$ (see Appendix \ref{app:ewc_deletion_objective}).
Therefore the information about $\mathcal{C}$ can be removed by scaling the current posterior $p(\boldtheta \given \mathcal{D})$ by the inverse of $p(\mathcal{C} \given \boldtheta)$ and re-normalising. 
In other words, we can treat the approximation of $p(\bm{\theta} \given \mathcal{D}\setminus\mathcal{C})$ as a continual learning task, where the task is to ``unlearn'' the datapoints in $\mathcal{C}$ while using the posterior distribution $p(\bm{\theta} \given \mathcal{D})$ as the prior. 
In practice, the target model to be fixed corresponds to the approximate posterior $q(\bm{\theta}) \approx p(\bm{\theta} \given \mathcal{D})$. Therefore continual (un)learning is done by
\begin{equation}\label{eq:approx_data_deletion}
    q\DmC(\bm{\theta}) \propto q(\bm{\theta}) / p(\mathcal{C} \given \boldtheta) \approx p(\boldtheta \given \mathcal{D}\setminus \mathcal{C}).
\end{equation}
The above approximation can be carried out with different approximate inference techniques such as MLE/MAP point estimate, Laplace approximation \cite{kirkpatrick2017overcoming} and variational inference \cite{nguyen2017variational}. Again we provide a few examples to concretise this process as follows. 

\textit{\textbf{Example 1 (Fine-tuning on Corrected Data): }}
Given a point estimate of model parameters $\hat{\boldtheta}$, i.e., $q(\boldtheta) = \delta(\boldtheta - \hat{\boldtheta})$, a simple way to approximate $p(\boldtheta \given \mathcal{D}\setminus \mathcal{C})$ is to fine-tune on the corrected dataset $\mathcal{D}\setminus \mathcal{C}$ and update the point estimate. The new $\hat{\boldtheta}\DmC$ of the repaired model are obtained by maximising the log-likelihood $\log \, p(\mathcal{D}\setminus\mathcal{C} \given \boldtheta)$ via SGD, starting from $\hat{\boldtheta}$. 

\textit{\textbf{Example 2 (Newton Update Removal): }}
Guo \etal \cite{guo2019certified} proposed a Newton update based method for data deletion. This method reduces to a specific form of Eq.~(\ref{eq:approx_data_deletion}) when using log-likelihood as its loss:
\begin{equation}\label{eq:koh_data_deletion}
\hat{\boldtheta}\DmC \approx \hat{\boldtheta} - \gamma \hat{\boldF}_{\hat{\boldtheta}}^{-1}\nabla_{\hat{\boldtheta}}\log \, p(\mathcal{C} \given \hat{\boldtheta}).
\end{equation}
Here information about $\mathcal{C}$ gets deleted by performing a single-step natural gradient descent on their log likelihood \cite{amari1998natural}. Also, notice the similarity with the way linear influence~\cite{koh2017understanding} is computed in Eq.~\eqref{eq:linear_influence}, illuminating the relation between cause identification and treatment steps.

\textit{\textbf{Example 3 (EWC for data deletion): }}
The update rule in Eq.~(\ref{eq:approx_data_deletion}) for data deletion is amenable to any continual learning approaches. For example, given model parameters $\hat{\boldtheta}$, EWC-based deletion obtains new parameters $\hat{\boldtheta}\DmC$ by maximising the following objective (see Appendix \ref{app:ewc_deletion_objective}):
\begin{equation}
     - \log \, p(\mathcal{C} \given \boldtheta)  - \frac{N}{2}(\boldtheta - \hat{\boldtheta})^\top  \hat{\boldF}_{\hat{\boldtheta}} (\boldtheta - \hat{\boldtheta}) - \frac{\lambda}{2}||\boldtheta - \hat{\boldtheta}||^2_2 
\label{eq:deletion_ewc}
\end{equation}
where the first term seeks to remove information about $\mathcal{C}$ while the remaining terms discourage parameters from deviating from the original values. Contrasting this with  Eq.~\eqref{eq:predictive_ewc} again reveals the connection between EWC methods for cause identification and treatment steps.

\blue{\textbf{Comparison:}
Similar to the comparison made in the cause identification part, EWC for data deletion also generalises the Newton update removal (Eq.~(\ref{eq:koh_data_deletion})). This can again be shown by deriving (damped) one-step fixed point iterative update starting from $\hat{\boldtheta}$ to approximate the fixed point of Eq.~(\ref{eq:deletion_ewc}) when $\lambda = 0$. As EWC for deletion uses SGD to approximate optimum of Eq.~(\ref{eq:deletion_ewc}), it is arguably better than Newton update removal for erasing the effects of detrimental examples $\mathcal{C}$, while better maintaining performance on other cases.}

\red{\textbf{Connecting the two steps:} We highlight the deep connection between the ``predictive approach'' for cause identification (Sec.~\ref{sec:cause_identification}), and the continual (un)learning for data deletion in the treatment step. They share the key idea of editing the approximate posterior $q(\bm{\theta})\approx p(\bm{\theta} \given \mathcal{D})$ via continual learning, which corresponds to editing the factor graph \citep{kschischang2001factor} of $p(\bm{\theta} | \mathcal{D})$, e.g., insertion of $p(\mathcal{F} | \bm{\theta})$ in cause identification, and deletion of $p(\mathcal{C}| \bm{\theta})$ in treatment. This unified view enables ones to take any continual learning method and extend it to both steps of model repairment. Indeed in our experiments, EWC as a better continual learning approach than e.g., one-step Newton update leads to improvement in both tasks over the prior works. }

% \YL{YL: if we have EWC results for repairment part, we can also write comparison between Newton update removal and EWC removal, similar to the discussion of connections between linear influence and EWC-influence.}

%%%%%%%%%%%%%%%%%%%%%%%%%%%%%%%%%%%%%%%%%%%%%%%%%%%%%%%%%%%%%%%%%%%%%%%%%
%\vspace{-3mm}
\section{Related work}
%\vspace{-3mm}
%
\textbf{Model Editing}.
% MEL: I find this paragraph still a little too long, I would shorten the beginning, the first sentence in original version seems a good one for conclusions.
There is a recent surge of interest in developing targeted updates to correct model's undesirable behaviours, while leaving other desired properties intact. As naive fine-tuning methods often lead to overfitting to the failure examples and accuracy degradation on others, various strategies have been proposed.
%As more and more large-scale models are deployed in real-world applications, the importance of maintaining their performance over time has become more evident. As a result, recent years have seen a surge in interest in the development of methods for making targeted updates to a model to correct its undesirable behaviours, while leaving it otherwise intact. Based on a finding that a naive fine-tuning on the failure examples with correct labels often leads to overfitting and compromises performance on other examples, various strategies have been proposed.
%
For example, Zhu \etal \cite{zhu2020modifyingmemories} employ a simple regularization technique to minimize parameter changes during the fine-tuning phase. Subsequent works \cite{sinitsin2020editable,cao2021editing} advocate for a functional regularisation instead, e.g. KL divergence in the output space, to achieve better regularisation. These lines of work, additionally, propose to use meta-learning \cite{finn2017model} to learn to edit the target model, where the latest meta-learning approach is proposed by Mitchell \etal \cite{mitchell2021fast}. Another promising approach \cite{santurkar2021editing} performs weight editing so that features of a specific concept (e.g. snow) map to the features of another (e.g. road). A commonality among these approaches is the focus on direct model edits for correction. Our work takes an orthogonal and under-explored angle where the aim is to ``edit'' the data instead, by identifying and removing harmful examples which cause failures --- in turn, this difference makes our framework complementary to these model-editing approaches. 

\textbf{Continual learning.} Continual learning is an active research area with a related but broader scope than model repairment, which aims to develop methods that adapt the model for future tasks while maintaining model performance on previously learned tasks \citep{parisi2019continual}.
We focus on a more targeted problem in this work, yet introduce a framework that allows the use of any continual learning approach for model repairment.
Our experiments presents EWC \citep{kirkpatrick2017overcoming} as a practical instantiation of the framework. One can also leverage improvements over EWC such as online EWC \citep{schwarz2018progress}, or other regularisation-based methods that are motivated by Bayesian learning principles, such as variational continual learning \citep{nguyen2017variational, pan2020continual, loo2021generalized}, synaptic intelligence \citep{zenke2017continual}, and orthogonal gradient descent \citep{farajtabar2020orthogonal}. As approximations to $r(\mathcal{C})$ rely on accurate posterior approximations, advances in Bayesian continual learning methods are expected to improve the practical effectiveness of model repairment under our framework.

\textbf{Data Selection and Valuation}.
Multiple techniques have been introduced for selecting ``influential'' training examples on a chosen metric (e.g. test accuracy), such as influence functions \citep{koh2017understanding,koh2019accuracy,barshan2020relatif,giordano2019swiss,hara2019data,khanna2019interpreting,warnecke2021machine}, Shapley value-based approaches \citep{ghorbani2019data,ancona2019explaining,jia2019towards} and probability of sufficiency \citep{chakarov2016debugging}.
Within the category of influence functions, two representative approaches include linear influence function \citep{koh2017understanding} and SGD-influence \citep{hara2019data}. The former approach performs one-step update only, thus, while efficient, it may be less accurate in reflecting the influence of a datum $\bm{z}$. The latter approach computes a projected difference between $\hat{\boldtheta}$ and $\hat{\boldtheta}\DF$ but with $\hat{\boldtheta}\DF$ obtained by running SGD fine-tuning on training data without $\bm{z}$. Thus SGD-influence is computationally inefficient. Compared to both baselines, our EWC-influence approach achieve the best in both worlds: it produces more accurate influence estimates than linear influence due to better optimisation, while it is more efficient than SGD-influence as it requires only one optimisation procedure on the given failure set $\mathcal{F}$.

% \begin{itemize}
%     %\item Influence Function: original paper, ICML'17 \citep{koh2017understanding}; batch-wise analysis, NeurIPS'19 \citep{koh2019accuracy}; local variant, AISTATS'20 \cite{barshan2020relatif}. Jacknife but related \citep{giordano2019swiss}?
%     %\item SGD-influence: NeurIPS'19 \citep{hara2019data}
%     %\item Fisher kernel: AISTATS'19 \citep{khanna2019interpreting}
%     %\item Shapley Value, ICML'19 \citep{ghorbani2019data} and its distributional extension, ICML'20 \citep{ghorbani2020distributional}
%     %\item Probability of Sufficiency, 2016, \citep{chakarov2016debugging}
%     % \item Measure of ``forgetable''-ness, ICLR'17 \cite{toneva2018empirical}
%     \item \textbf{Other applications (Optional?)}: \YL{YL: need to decide whether they are relavent}
%     \begin{enumerate}
%         \item Robust learning by re-weighting training data based on a small clean dataset e.g. MentorNet \citep{jiang2018mentornet}, Meta-learn to reweight \citep{ren2018learning}, Reinforcement Learning \citep{yoon2019data}, Sample re-weighting: heteroscedastic models (refs), gradient variance \cite{katharopoulos2018not}, meta-learning \cite{ren2018learning}, a similar idea but uses reinforcement learning \cite{yoon2019data}
%         \item Summarisation (e.g. coreset)
%         \item Active learning, Bayes Optimisation
%     \end{enumerate}
% \end{itemize}

% The above methods have indeed shown that selecting ``bad'' training datapoints based on these criteria, removing them and re-training the model leads to an improvement in performance. 

\textbf{Data Deletion}.
The detrimental data removal in the treatment step is related to \emph{data deletion}, a rapidly developing field of machine learning research \citep{bourtoule2019machine,guo2019certified,ginart2019making,izzo2020approximate,neel2020descent,gupta2021adaptive}. 
Closest to our work is variational Bayesian unlearning \citep{nguyen2020variational} which extends variational Bayes to data deletion settings. But the connection to continual learning is not explicitly made, and it is limited to applications in logistic regression and sparse Gaussian processes.
In general, the main focus of existing data deletion research is to preserve data privacy, and datapoints to be removed are assumed provided. On the contrary, in this work, we focus on the repairment of models and propose a unified procedure not only to remove data but also to identify which ones to do so. 
%In general, the main focus of existing datadeletion research is to preserve data privacy, as datapoints tobe removed need to be provided to the deletion algorithms. Onthe contrary, in the context of model repairment, we assumeaccess to data during repairment and emphasise on a unifiedprocedure of detrimental data identification and remova

%%%%%%%%%%%%%%%%%%%%%%%%%%%%%%%%%%%%%%%%%%%%%%%%%%%%%%%%%%%%%%%%%%%%%%%%%%%%%%%

%\vspace{-2mm}
\section{Experiments}
%\vspace{-3mm}
We evaluate the efficacy of the proposed framework in a) identifying the causes of target prediction failures in Sec.~\ref{sec:results_causes}, and b) repairing the original model by erasing the memories of such causes in Sec.~\ref{sec:results_treatment}. We use augmented versions of MNIST and CIFAR-10 datasets with simulated annotation and input noise. Such controlled experiments are performed for creating "ground truths” of failure causes -- necessary for validating the quality of identification methods -- and for testing the method in a variety of settings. 

\textbf{Baselines.} For the cause identification task, we compare our approach (\textit{EWC-influence}) against the linear influence function \cite{koh2017understanding} and \textit{SGD-influence} \cite{hara2019data}. To avoid expensive computation of $\hat{\boldF}_{\hat{\boldtheta}}^{-1}$, Koh $\&$ Liang \cite{koh2017understanding} introduced two efficient approximations to the Hessian-vector product $\hat{\boldF}_{\hat{\boldtheta}}^{-1} \nabla_{\boldtheta} \log p(\mathcal{F} \vert\boldtheta)$; the first solves $\text{arg min}_{v}\{v^{T}\hat{\boldF}_{\hat{\boldtheta}}^{-1} \nabla_{\boldtheta}v - \log p(\mathcal{F} \vert\boldtheta)^{T}v\}$ with gradient descent (GD), while the second uses an iterative algorithm for stochastic approximation (SA) from \citep{agarwal2016second}.
%---we henceforth refer to them as \textit{linear influence (Gradient Descent or GD in short)} and \textit{linear influence (Stochastic Approximation or SA in short)}. 
We implement these two variants (GD \& SA) of linear influence in Pytorch, and use the original implementation for SGD-influence. For the model treatment task, we compare our method (\textit{EWC-deletion}) against Newton update removal \cite{guo2019certified}. This method again requires computing a Hessian-vector product for which we employ the same stochastic approximation technique \cite{agarwal2016second}. To isolate the evaluation of cause identification and treatment, we further consider in Section \ref{sec:results_causes} \emph{fine-tuning} on $\mathcal{D}\setminus \mathcal{C}$ as another repairment strategy, which would return the best repairment result if the set $\mathcal{C}$ correctly captures the detrimental datapoints. Lastly, we set the prior term $\lambda$ to zero in eq.\eqref{eq:predictive_ewc} and eq.~\eqref{eq:deletion_ewc} to ensure fair comparison with linear influence and Newton update removal.  
%Lastly for fair comparision with prior works on linear influence eq.(11) and Newton update removal eq.(17), we set the prior term $\lambda$ to zero in our experiments.

\textbf{Common Set-up.} We train the base classification models on the training split of the ``augmented'' MNIST and CIFAR-10 datasets.
For MNIST, we use $6\%$ (3000 samples) of the original training set to make the task more challenging. We use instances of CNNs throughout and train them using the Adam optimiser \cite{kingma2014adam}. The architecture and training details can be found in Appendix \ref{sec:supp_experimental_details}. 
%For experiments on both cause identification and treatment, once the base models are trained, 
For evaluation, we separate the test set $\mathcal{T}$ into the set of misclassified examples, $\mathcal{F}$ (``\textit{failure set}'') and the others, $\mathcal{T}\setminus\mathcal{F}$ which are correctly classified (``\textit{remaining set}''). We further split the failure set into \textit{query}, $\mathcal{F}_{q}$ and \textit{holdout}, $\mathcal{F}_{h}$ sets, where we only use the former to identify failure causes $\mathcal{C}$, and use the latter to quantify how generalisably the removal of $\mathcal{C}$ can amend the failure cases. We stress that $\mathcal{F}_q$ is used for cause identification only, but not for further model adaptation. % in contrast with prior works such as \cite{sinitsin2020editable}. 

\begin{figure*}[t]
\centering
%\vspace{-16mm}
  \includegraphics[width=1.0\linewidth]{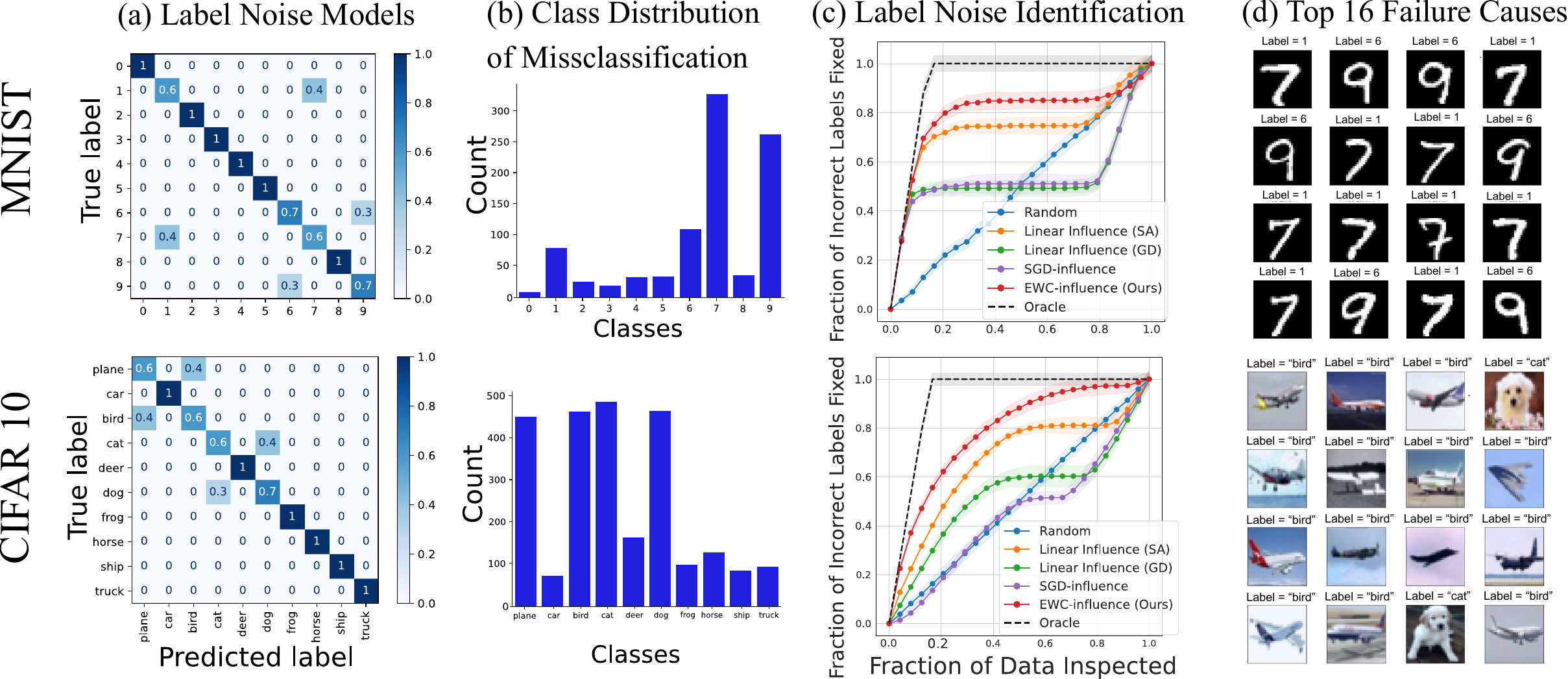}
  \vspace{-4mm}
  \caption{\footnotesize %{\bf EWC-influence outperforms other baselines at cause identification in the presence of annotation noise.}
  Results on cause identification in the presence of annotation noise.
  (a) shows the confusion matrices used to simulate class-dependent label noise on MNIST and CIFAR-10. (b) shows the class distribution of the misclassified examples for a single run. (c) plots how much of the  identified causes match the samples with incorrect labels for different approaches. The shade represents the standard deviation computed from 5 different runs. (d) shows the top 16 causes of the failures as ranked by EWC-influence. An enlarged version can be found in Appendix~\ref{sec:supp_enlarged_figs}.}
  \label{fig:cause-identification}
  %\caption{\footnotesize Results for cause identification in the presence of annotation noise. (a) describes the confusion matrices, based on which class-dependent label noise are simulated on MNIST and CIFAR10 datasets. (b) quantifies how closely the identified causes match the samples with annotation noise for the respective approaches. The shade represents the standard deviation computed from 5 different runs. (c) shows the class distributions of the failure cases for one of the runs, used in the cause identification step. Lastly, (d) shows the corresponding 16 strongest causes of the failures according to \textit{ewc-influence}, most of which coincide with examples with wrong labels. }
  \label{fig:labelnose_detection}
  \vspace{-2mm}
\end{figure*}

\begin{figure}[t]
\centering
  \includegraphics[width=1.0\linewidth]{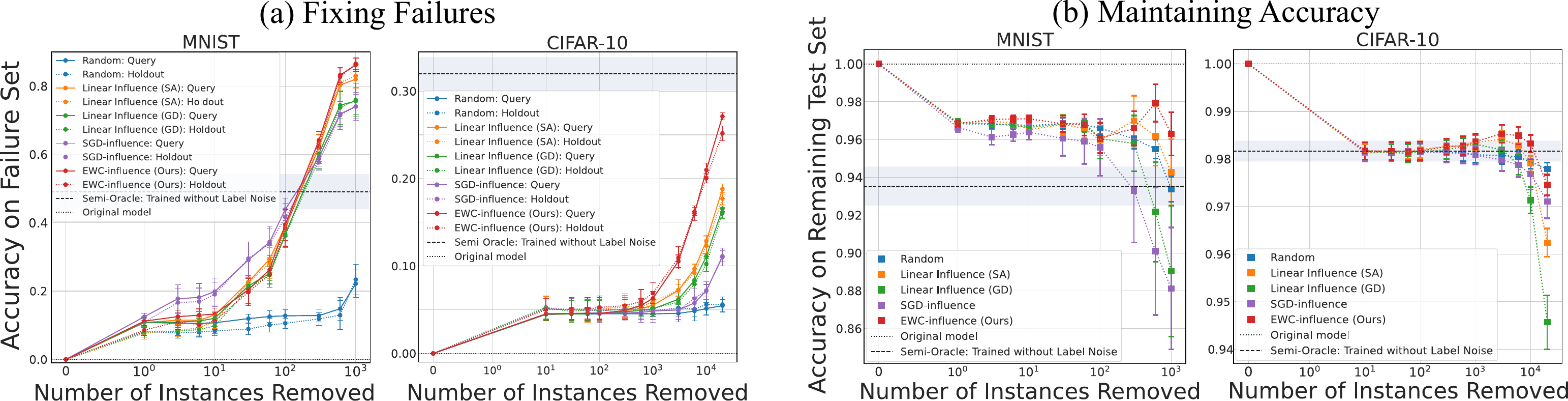}
  \vspace{-3mm}
  \caption{\footnotesize Comparison of the quality of identified causes in the presence of annotation noise. The impact of gradually removing samples in MNIST and CIFAR-10 datasets in the order of influence values $r(\bz)$ are measured on the failure sets (holdout and query) in (a), and on the remaining test set in (b). \blue{We note that in (b), accuracy values start at 1.0 as they are calculated on the set of test samples on which the original model makes correct predictions}. We also plot the performance of another reference (``semi-oracle'') that is the original model fine-tuned on the training data without the label noise instances. The means/stds of all quantities are calculated over 5 runs. See Appendix~\ref{sec:supp_enlarged_figs} for an enlarged version.}
  \label{fig:causes_quality}
  \vspace{-1mm}
\end{figure}

%\vspace{-4mm}
\subsection{Identifying Failure Causes}\label{sec:results_causes}
%\vspace{-6mm}
% MEL: repeated, delete
%In the first series of experiments, we evaluate how well the proposed EWC-influence is able to identify causes of failures under different settings. 

\begin{figure*}[t]
\centering
%\vspace{-17mm}
  \includegraphics[width=1.0\linewidth]{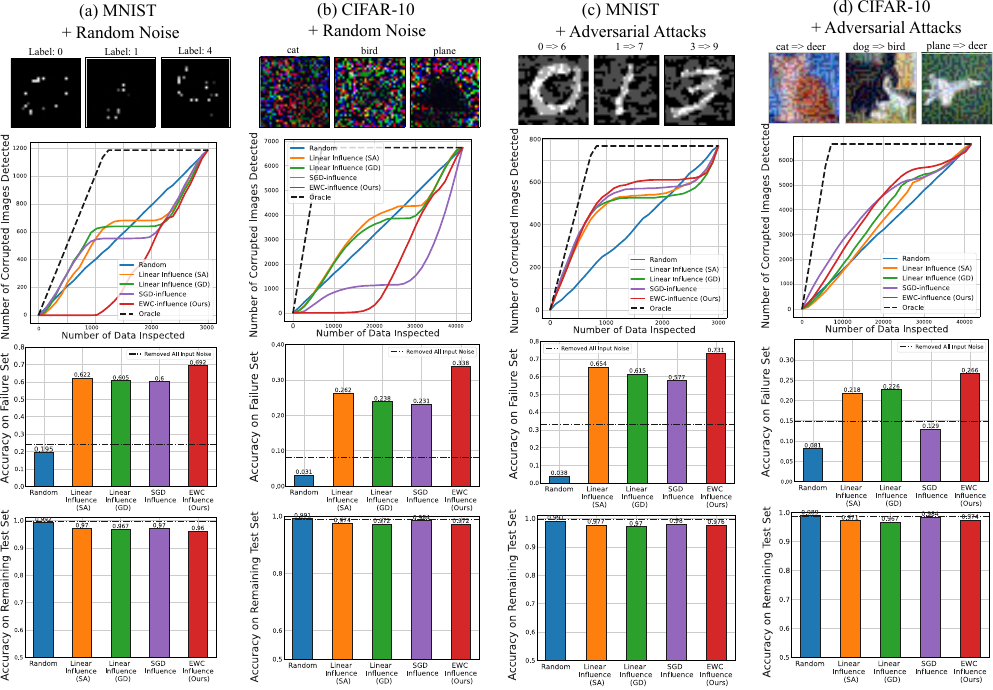}
  \vspace{-5mm}
  \caption{\footnotesize 
  %{\bf EWC-Influence identifies successfully "harmful" (adversarial) input noise while avoiding harmless (random) input noise.}
  Results on cause identification in the presence of different input noises. From top to bottom, we show i) examples of corrupted samples (synthetic proxy for potential causes of failure), ii) how many of the identified causes correspond to samples corrupted with input noise, iii) and iv) performance in failure holdout set $\mathcal{F}_{h}$ and remaining test set when removing the top 1000/20000 identified causes in MNIST/CIFAR-10. The influence values are calculated with respect to 50\% of test-time failure cases that belong to the classes that suffer from input noise. EWC-Influence identifies ``harmful'' (adversarial) input noise better than random while avoiding ``harmless'' (random) input noise. An enlarged version is available in Appendix~\ref{sec:supp_enlarged_figs}.
  %\MFP{(Would it make sense to highlight in 2nd row the cuting plane, i.e., how many removed data does 3rd and 4th row correspond to?)}  
  %(a,b)  Random input noise for MNIST (top row) and CIFAR10 (bottom row) datasets. 
  }
  \label{fig:input_noise_detection}
   \vspace{-4mm}
\end{figure*}

% MEL: Consider calling this label noise, or at least mention it. It would be more precise
\paragraph{Annotation Noise.}
%Figure~\ref{fig:cause-identification} explores how well the proposed EWC-influence approach is able to identify causes of failure.
To induce test prediction failures, we randomly flip labels in the training set between semantically similar classes (e.g. 1 and 7 for MNIST, and cats and dogs for CIFAR-10) according to the confusion matrices in Fig.~\ref{fig:cause-identification}(a). As a result, the classes of miss-classified test examples are concentrated on those classes with label noise as depicted in Fig.~\ref{fig:cause-identification}(b).  %

To measure the accuracy of identifying incorrectly labelled examples, we inspect the training examples $\bz \in \mathcal{D}$ in the descending order of $\tilde{r}(\bz)$ computed with $\mathcal{F}_q$ which contains 50\% of the miss-classified test cases, and calculate the fraction of incorrectly labelled datapoints in inspected examples. Fig.~\ref{fig:cause-identification}(c) shows that EWC-influence identifies more failure causes earlier on compare to other methods, and is the closest match to the ``Oracle'' baseline which has full knowledge of samples with wrong labels. Fig.~\ref{fig:cause-identification}(d) shows that the top few causes according to EWC-influence are the samples with incorrect labels, while the least harmful ones are the images of the same classes but with the correct labels as shown in Fig.~\ref{fig:label_noise_least_harmful} in Appendix \ref{sec:supp_extra_results}. 

%Next, we seek to quantify the \emph{quality of causes} identified by the respective influence methods. While annotation noise is one of the detrimental factors, the prediction failures may also arise from other types of harmful examples. 
%

As stipulated in Sec.~\ref{sec:method}, a set of identified causes $\mathcal{C}$ is of higher quality if removing them leads to a larger gain in accuracy on the failure set while maintaining performance on others. To measure such \emph{quality of causes}, we fine-tune the base model on $\mathcal{D}\setminus \mathcal{C}$
and report the accuracy on the failure query set $\mathcal{F}_{q}$, the holdout failure set $\mathcal{F}_{h}$ as well as the remaining test set, $\mathcal{T}\setminus\mathcal{F}$. Results in Fig.~\ref{fig:causes_quality} suggest that removing failure causes according to EWC-influence yields the highest increase in accuracy on the failure set $\mathcal{F}$ without hurting performance in the remaining test set $\mathcal{T} \setminus \mathcal{F}$. We also note that all of the methods are able to fix the failures better than randomly removing datapoints, and more interestingly, for MNIST, when enough causes are erased ($\approx 10^3$), all methods even surpass the case in which all label noise instances are removed. This result implies that, while annotation noise is a major detrimental factor, the prediction failures also arise from other types of harmful examples.

Lastly, we evaluate the sample efficiency of cause identification by reducing the size of the query set $\mathcal{F}_{q}$.  Fig.~\ref{fig:sample_efficiency} in Appendix \ref{sec:supp_extra_results} shows that all approaches degrade gracefully in repairment performance as the query size gets smaller, but overall EWC-influence still remains the best in terms of label noise detection and repairment accuracy on the failure set and the remaining set. %The biggest drop occurs in Accuracy on the remaining holdout set: EWC-influence presents a stronger degradation in small sample regimes (around 10\% of the original failure set) but without dropping below other approaches. 

% Previous version
% Figure~\ref{fig:sample_efficiency} compares the sample efficiency of different approaches for cause identification on MNIST and CIFAR10 datasets with annotation noise. More specifically, we want to understand how many failure cases in the failure set \mathcal{F} we need to see for cause identification to work at all, i.e., how performance of each approach varies as we reduce \MFP{It might be more intuitive if we were to plot mirrored x-axis, such that left side corresponds to original query set (assuming that is the size used for Fig. 2.)} the size of the failure set. EWC-influence approach remains the winner in terms of precision, recall, and accuracy on failure set even as we decrease the size of the failure (query) set. All approaches exhibit a stable behavior w.r.t size of the failure set. The biggest drop occurs in Accuracy on the remaining holdout set: EWC-influence presents a stronger degradation in small sample regimes (around 10\% of the original failure set) but without dropping below other approaches. Worth mentioning, note that for MNIST, all approaches improves above the semi-oracle that trains with an uncorrupted training set: this means that all methods not only correct the synthetically added examples, but they also remove other harmful examples that are naturally present to begin with.

\textbf{Random Input Noise.}
In this experiment, we inject synthetic outliers into MNIST and CIFAR-10 and test the quality of cause identification. We select a set of target classes~---~1, 7, 6, 9 for MNIST and plane, bird, cat, dog for CIFAR-10~---~and randomly corrupt 30\% of the images in those classes by adding salt-and-pepper noise \blue{(i.e. replacing pixels with extreme values 0 and 255)} in MNIST and Gaussian noise in CIFAR-10. The top row in Fig.~\ref{fig:input_noise_detection} shows examples, and those corrupted images constitute roughly 12\% of the whole training set. Sec.~\ref{sec:supp_experimental_details} in the appendix provides details. 

We use a subgroup of failures in the target classes as the query $\mathcal{F}_q$ to compute influence values. Surprisingly, the second rows in Fig.~\ref{fig:input_noise_detection}(a) and (b) show that EWC-influence largely avoids selecting the corrupted images as the top 1000 causes for MNIST and the top 20000 causes for CIFAR-10. However, the third and the fourth rows show that removing those causes results in the best treatment performance on failures while maintaining the performance at a level similar to other baselines. In fact, removing all the input noise and retraining is not able to fix the failures by much, indicating that EWC-influence is able to correctly avoid these relatively harmless outliers and detect other more harmful causes. Fig.~\ref{fig:input_not_harmful} in Appendix \ref{sec:supp_extra_results} visualises the most harmful examples identified by EWC-influence. 
%First of all, most of these examples are from the non-target classes. While the input noise itself may not influence the training of the model, the sample size of clean images in the target classes is still smaller---such group imbalance can be rectified by subsampling the dominant group as implicated by this result. Secondly, 
Many of them appear to be ambiguous instances in non-target classes, e.g. wonky digits, close-up views of vehicles, a real instance with incorrect label \cite{northcutt2021confidentlearning}, etc.

\textbf{Adversarial Poisoning.}
To simulate input noise that can induce test-time failures, we introduce contaminated data by randomly corrupting 30\% of the training images in those previously mentioned target classes. These poisoned datapoints are adversarial images crafted by the fast gradient sign method (FGSM)~\citep{goodfellow2014explaining} on a separate set of victim models trained on the original clean datasets, and they are labelled by the classes predicted by the victim models. The poisoned datasets are then used to train the base models that are used for evaluation of cause identification. Fig.~\ref{fig:input_noise_detection}(c) and (d) show that most of the influence functions detect the corrupted samples better than the ``random'' baseline. The dashed lines in the third row show that removing all of the corrupted inputs lead to a significant gain in accuracy on the holdout failure set in comparison with the random noise setting, illustrating the larger extent of harms caused by data poisoning. However, most of the identification methods still outperform this reference by a large margin. This suggests again for the presence of other influential samples, and EWC-influence is able to pick up the most important ones, judging by the accuracy on the failure set.

\textbf{Speed Comparison.} Table~\ref{tab:computation} in Appendix \ref{sec:supp_extra_results} shows the total run-time of cause identification methods on a single GPU for their best sets of hyper-parameters selected based on the treatment accuracy on the failure set. For both datasets, EWC-influence achieves comparable or shorter run time than the baselines. 

%\paragraph{Label Noise}
%See Fig.~1. 
%\begin{itemize}
%    \item Label noise for classes (precision & recall curves) 
%    \item Input noise for specific digits (precision & recall curves)
%    \item Comparison of different identification methods with the Oracle adaptation %method/fine-tuning. 
%    \item Sample efficiency results
%\end{itemize}
%\RT{3 identification methods to add: multi-step linear influence, linear influence with diagonal hessian, EWC with laplace approximation}

% Fig.2 in Data Shapley \cite{ghorbani2019data}
% Robustness of the identified cases. 

%\paragraph{Input Noise}
%\begin{itemize}
%    \item Random noise (Fig.~5) 
%    \item Adversarial Data (coming soon)
%\end{itemize}

%\vspace{-3mm}
\subsection{Comparison of Treatment Methods}\label{sec:results_treatment}
\begin{figure}[t!]
\centering
   \vspace{-2mm}
  \includegraphics[width=0.6\linewidth]{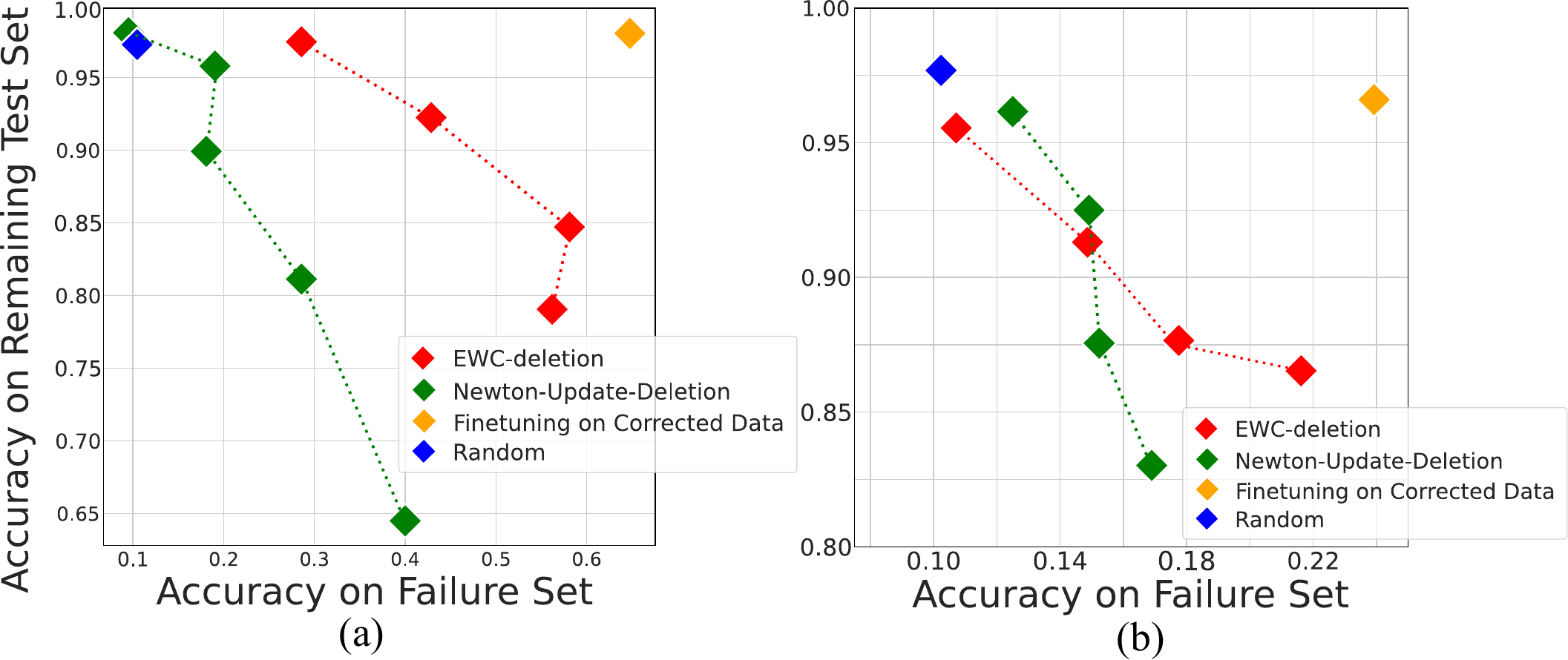}
    \vspace{-3mm}
  \caption{\footnotesize Comparison of deletion-based treatment methods on (a) MNIST and (b) CIFAR-10. For Newton-update-deletion and EWC-deletion, we plot multiple results for varying hyper-parameters to visualise the trade-off between the accuracy on the failure set and the remaining set. The closer to the top right corner, the more desirable.}
    \vspace{-3mm}

  \label{fig:deletion_comparison}
\end{figure}

%\vspace{-3mm}
We evaluate the performance of different deletion-based methods for treatment introduced in Sec.~\ref{sec:treatment} on MNIST and CIFAR10 datasets with simulated annotation noise, used in the previous section. %For EWC-deletion (our method), we vary the amount of weight regularisation --- the second term in Eq.~\eqref{eq:deletion_ewc} --- by scaling it by $2/\rho n$ where $n$ denotes the number of training datapoints, and $\rho>0$ is the new hyperparameter. For Newton-update deletion \cite{guo2019certified}, we also vary the step size of the gradient ascent by scaling the second term in Eq.~\eqref{eq:koh_data_deletion} by $\rho>0$. We run both methods for different values of $\rho$ in the range $[0.01, 0.05]$ until convergence with early stopping based on the query set accuracy. 
We run both EWC-deletion (ours) and Newton update removal \citep{guo2019certified} methods with early stopping based on the query set accuracy, and experiment with different hyper-parameter settings (see Sec.~\ref{sec:supp_experimental_details} in Appendix) to achieve different trade-offs between failure set accuracy and remaining set performance. Here EWC-influence is used to identify the causes, and the top 15\% examples were removed by the respective deletion methods. 
Such trade-off is shown in Fig.~\ref{fig:deletion_comparison}, where fine-tuning on $\mathcal{D}\setminus\mathcal{C}$ is included as an ``upper-bound'' reference for data deletion performance. %\footnote{
%The two data deletion methods require only access to the causes $\mathcal{C}$ and are therefore much faster than fine-tuning.}
%
On MNIST, EWC-deletion attains a considerably better trade-off between treatment and maintenance compared to Newton-update-deletion, and is much closer to the fine-tuning reference. For CIFAR-10, EWC-deletion beats the Newton-update deletion by 5\% in the best failure accuracy while the order reverses for the best accuracy on the remaining test set but with less than 1\% difference.

\section{Conclusions}
%\vspace{-4mm}

In this work, we develop a generic framework for \textit{repairing} machine learning models by erasing memories of detrimental datapoints. The framework consists of two key components, that are, the mechanism for identifying the ``causes'' in training data which are responsible for the given failures, and the adaptation method for fixing the model by removing information about them. The two components are connected under the Bayesian view of continual (un)learning, which brings forth several practical benefits. Firstly, the framework subsumes some recent works on influence function and data deletion as specific examples, and elucidate their limitations. Secondly, the generality of our approach allows leveraging recent advances in continual learning in this new problem of model repairment. In particular, we extend Elastic Weight Consolidation to cause identification and data deletion, and demonstrate empirically its competitive performance in both tasks.
% Future work will investigate the values of adapting more recent continual learning approaches, and also study 
% MEL: study -> combine with? %other types of data correction mechanisms (e.g. label correction, sample/label acquisition). 

\section*{Acknowledgements}
We would like to thank Ozan Oktay, Stephanie Hyland, and Ted Meeds at Microsoft Research Cambridge and Jin Chen at University College London for their valuable feedback on an early version of this work. 
% \RT{\lipsum[66] \lipsum[65] }

% MEL: Material for COnclusion
% Emphasize: Even when datasets are curated conscientiously before training, unexpected failures might still occur at deployment, compromising safety and incurring high costs. 

% \section*{Broader Impact}
\bibliographystyle{unsrt}
% \bibliographystyle{iclr2021_conference}
% \bibliographystyle{icml2021}
%\input{main.bbl}
%\small
\bibliography{main}

\begin{thebibliography}{10}

\bibitem{frenay2013classification}
Beno{\^\i}t Fr{\'e}nay and Michel Verleysen.
\newblock Classification in the presence of label noise: a survey.
\newblock {\em IEEE transactions on neural networks and learning systems},
  25(5):845--869, 2013.

\bibitem{xu2020adversarial}
Han Xu, Yao Ma, Hao-Chen Liu, Debayan Deb, Hui Liu, Ji-Liang Tang, and Anil~K
  Jain.
\newblock Adversarial attacks and defenses in images, graphs and text: A
  review.
\newblock {\em International Journal of Automation and Computing},
  17(2):151--178, 2020.

\bibitem{koh2021wilds}
Pang~Wei Koh, Shiori Sagawa, Sang~Michael Xie, Marvin Zhang, Akshay
  Balsubramani, Weihua Hu, Michihiro Yasunaga, Richard~Lanas Phillips, Irena
  Gao, Tony Lee, et~al.
\newblock Wilds: A benchmark of in-the-wild distribution shifts.
\newblock In {\em International Conference on Machine Learning}, pages
  5637--5664. PMLR, 2021.

\bibitem{torralba2011unbiased}
Antonio Torralba and Alexei~A Efros.
\newblock Unbiased look at dataset bias.
\newblock In {\em CVPR 2011}, pages 1521--1528. IEEE, 2011.

\bibitem{chen2018my}
Irene Chen, Fredrik~D Johansson, and David Sontag.
\newblock Why is my classifier discriminatory?
\newblock In {\em Advances in Neural Information Processing Systems}, pages
  3539--3550, 2018.

\bibitem{sagawa2019distributionally}
Shiori Sagawa, Pang~Wei Koh, Tatsunori~B Hashimoto, and Percy Liang.
\newblock Distributionally robust neural networks for group shifts: On the
  importance of regularization for worst-case generalization.
\newblock {\em arXiv preprint arXiv:1911.08731}, 2019.

\bibitem{nguyen2017variational}
Cuong~V Nguyen, Yingzhen Li, Thang~D Bui, and Richard~E Turner.
\newblock Variational continual learning.
\newblock {\em arXiv preprint arXiv:1710.10628}, 2017.

\bibitem{koh2017understanding}
Pang~Wei Koh and Percy Liang.
\newblock Understanding black-box predictions via influence functions.
\newblock In {\em ICML}, 2017.

\bibitem{guo2019certified}
Chuan Guo, Tom Goldstein, Awni Hannun, and Laurens van~der Maaten.
\newblock Certified data removal from machine learning models.
\newblock {\em arXiv preprint arXiv:1911.03030}, 2019.

\bibitem{kirkpatrick2017overcoming}
James Kirkpatrick, Razvan Pascanu, Neil Rabinowitz, Joel Veness, Guillaume
  Desjardins, Andrei~A Rusu, Kieran Milan, John Quan, Tiago Ramalho, Agnieszka
  Grabska-Barwinska, et~al.
\newblock Overcoming catastrophic forgetting in neural networks.
\newblock {\em Proceedings of the national academy of sciences},
  114(13):3521--3526, 2017.

\bibitem{henn2021principled}
Thomas Henn, Yasukazu Sakamoto, Cl{\'e}ment Jacquet, Shunsuke Yoshizawa,
  Masamichi Andou, Stephen Tchen, Ryosuke Saga, Hiroyuki Ishihara, Katsuhiko
  Shimizu, Yingzhen Li, et~al.
\newblock A principled approach to failure analysis and model repairment:
  Demonstration in medical imaging.
\newblock In {\em International Conference on Medical Image Computing and
  Computer-Assisted Intervention}, pages 509--518. Springer, 2021.

\bibitem{parisi2019continual}
German~I Parisi, Ronald Kemker, Jose~L Part, Christopher Kanan, and Stefan
  Wermter.
\newblock Continual lifelong learning with neural networks: A review.
\newblock {\em Neural Networks}, 113:54--71, 2019.

\bibitem{mackay1992bayesian}
David~JC MacKay.
\newblock {\em Bayesian methods for adaptive models}.
\newblock PhD thesis, California Institute of Technology, 1992.

\bibitem{jordan1998introduction}
Michael~I Jordan, Zoubin Ghahramani, Tommi~S Jaakkola, and Lawrence~K Saul.
\newblock An introduction to variational methods for graphical models.
\newblock In {\em Learning in graphical models}, pages 105--161. Springer,
  1998.

\bibitem{blundell2015weight}
Charles Blundell, Julien Cornebise, Koray Kavukcuoglu, and Daan Wierstra.
\newblock Weight uncertainty in neural networks.
\newblock In {\em International conference on machine learning}, pages
  1613--1622. PMLR, 2015.

\bibitem{amari1998natural}
Shun-Ichi Amari.
\newblock Natural gradient works efficiently in learning.
\newblock {\em Neural computation}, 10(2):251--276, 1998.

\bibitem{kschischang2001factor}
Frank~R Kschischang, Brendan~J Frey, and H-A Loeliger.
\newblock Factor graphs and the sum-product algorithm.
\newblock {\em IEEE Transactions on information theory}, 47(2):498--519, 2001.

\bibitem{zhu2020modifyingmemories}
Chen Zhu, Ankit Singh~Rawat, Manzil Zaheer, Srinadh Bhojanapalli, Daliang Li,
  Felix Yu, and Sanjiv Kumar.
\newblock Modifying memories in transformer models.
\newblock {\em arXiv preprint arXiv:2012.00363}, 2020.

\bibitem{sinitsin2020editable}
Anton Sinitsin, Vsevolod Plokhotnyuk, Sergei Popov, and Artem Babenko.
\newblock Editable neural networks.
\newblock {\em arXiv preprint arXiv:2004.00345}, 2020.

\bibitem{cao2021editing}
Nicola~De Cao, Wilker Aziz, and Ivan Titov.
\newblock Editing factual knowledge in language models.
\newblock 2021.

\bibitem{finn2017model}
Chelsea Finn, Pieter Abbeel, and Sergey Levine.
\newblock Model-agnostic meta-learning for fast adaptation of deep networks.
\newblock In {\em International Conference on Machine Learning}, pages
  1126--1135. PMLR, 2017.

\bibitem{mitchell2021fast}
Eric Mitchell, Charles Lin, Antoine Bosselut, Chelsea Finn, and Christopher~D.
  Manning.
\newblock Fast model editing at scale.
\newblock {\em CoRR}, 2021.

\bibitem{santurkar2021editing}
Shibani Santurkar, Dimitris Tsipras, Mahalaxmi Elango, David Bau, Antonio
  Torralba, and Aleksander Madry.
\newblock Editing a classifier by rewriting its prediction rules.
\newblock In {\em Neural Information Processing Systems (NeurIPS)}, 2021.

\bibitem{schwarz2018progress}
Jonathan Schwarz, Wojciech Czarnecki, Jelena Luketina, Agnieszka
  Grabska-Barwinska, Yee~Whye Teh, Razvan Pascanu, and Raia Hadsell.
\newblock Progress \& compress: A scalable framework for continual learning.
\newblock In {\em International Conference on Machine Learning}, pages
  4528--4537. PMLR, 2018.

\bibitem{pan2020continual}
Pingbo Pan, Siddharth Swaroop, Alexander Immer, Runa Eschenhagen, Richard~E
  Turner, and Mohammad~Emtiyaz Khan.
\newblock Continual deep learning by functional regularisation of memorable
  past.
\newblock In {\em Advances in Neural Processing Information Systems}, 2020.

\bibitem{loo2021generalized}
Noel Loo, Siddharth Swaroop, and Richard~E Turner.
\newblock Generalized variational continual learning.
\newblock In {\em International Conference on Learning Representations}, 2021.

\bibitem{zenke2017continual}
Friedemann Zenke, Ben Poole, and Surya Ganguli.
\newblock Continual learning through synaptic intelligence.
\newblock In {\em International Conference on Machine Learning}, pages
  3987--3995. PMLR, 2017.

\bibitem{farajtabar2020orthogonal}
Mehrdad Farajtabar, Navid Azizan, Alex Mott, and Ang Li.
\newblock Orthogonal gradient descent for continual learning.
\newblock In {\em International Conference on Artificial Intelligence and
  Statistics}, pages 3762--3773. PMLR, 2020.

\bibitem{koh2019accuracy}
Pang Wei~W Koh, Kai-Siang Ang, Hubert Teo, and Percy~S Liang.
\newblock On the accuracy of influence functions for measuring group effects.
\newblock In {\em Advances in Neural Information Processing Systems}, pages
  5254--5264, 2019.

\bibitem{barshan2020relatif}
Elnaz Barshan, Marc-Etienne Brunet, and Gintare~Karolina Dziugaite.
\newblock Relatif: Identifying explanatory training examples via relative
  influence.
\newblock {\em arXiv preprint arXiv:2003.11630}, 2020.

\bibitem{giordano2019swiss}
Ryan Giordano, William Stephenson, Runjing Liu, Michael Jordan, and Tamara
  Broderick.
\newblock A swiss army infinitesimal jackknife.
\newblock In {\em The 22nd International Conference on Artificial Intelligence
  and Statistics}, pages 1139--1147, 2019.

\bibitem{hara2019data}
Satoshi Hara, Atsushi Nitanda, and Takanori Maehara.
\newblock Data cleansing for models trained with sgd.
\newblock In {\em Advances in Neural Information Processing Systems}, pages
  4213--4222, 2019.

\bibitem{khanna2019interpreting}
Rajiv Khanna, Been Kim, Joydeep Ghosh, and Sanmi Koyejo.
\newblock Interpreting black box predictions using fisher kernels.
\newblock In {\em The 22nd International Conference on Artificial Intelligence
  and Statistics}, pages 3382--3390, 2019.

\bibitem{warnecke2021machine}
Alexander Warnecke, Lukas Pirch, Christian Wressnegger, and Konrad Rieck.
\newblock Machine unlearning of features and labels.
\newblock {\em arXiv preprint arXiv:2108.11577}, 2021.

\bibitem{ghorbani2019data}
Amirata Ghorbani and James Zou.
\newblock Data shapley: Equitable valuation of data for machine learning.
\newblock In {\em ICML}, 2019.

\bibitem{ancona2019explaining}
Marco Ancona, Cengiz Oztireli, and Markus Gross.
\newblock Explaining deep neural networks with a polynomial time algorithm for
  shapley value approximation.
\newblock In {\em International Conference on Machine Learning}, pages
  272--281. PMLR, 2019.

\bibitem{jia2019towards}
Ruoxi Jia, David Dao, Boxin Wang, Frances~Ann Hubis, Nick Hynes, Nezihe~Merve
  G{\"u}rel, Bo~Li, Ce~Zhang, Dawn Song, and Costas~J Spanos.
\newblock Towards efficient data valuation based on the shapley value.
\newblock In {\em The 22nd International Conference on Artificial Intelligence
  and Statistics}, pages 1167--1176. PMLR, 2019.

\bibitem{chakarov2016debugging}
Aleksandar Chakarov, Aditya Nori, Sriram Rajamani, Shayak Sen, and Deepak
  Vijaykeerthy.
\newblock Debugging machine learning tasks.
\newblock {\em arXiv preprint arXiv:1603.07292}, 2016.

\bibitem{bourtoule2019machine}
Lucas Bourtoule, Varun Chandrasekaran, Christopher Choquette-Choo, Hengrui Jia,
  Adelin Travers, Baiwu Zhang, David Lie, and Nicolas Papernot.
\newblock Machine unlearning.
\newblock {\em arXiv preprint arXiv:1912.03817}, 2019.

\bibitem{ginart2019making}
Antonio Ginart, Melody Guan, Gregory Valiant, and James~Y Zou.
\newblock Making ai forget you: Data deletion in machine learning.
\newblock In {\em Advances in Neural Information Processing Systems}, pages
  3518--3531, 2019.

\bibitem{izzo2020approximate}
Zachary Izzo, Mary~Anne Smart, Kamalika Chaudhuri, and James Zou.
\newblock Approximate data deletion from machine learning models: Algorithms
  and evaluations.
\newblock {\em arXiv preprint arXiv:2002.10077}, 2020.

\bibitem{neel2020descent}
Seth Neel, Aaron Roth, and Saeed Sharifi-Malvajerdi.
\newblock Descent-to-delete: Gradient-based methods for machine unlearning.
\newblock {\em arXiv preprint arXiv:2007.02923}, 2020.

\bibitem{gupta2021adaptive}
Varun Gupta, Christopher Jung, Seth Neel, Aaron Roth, Saeed Sharifi-Malvajerdi,
  and Chris Waites.
\newblock Adaptive machine unlearning.
\newblock {\em arXiv preprint arXiv:2106.04378}, 2021.

\bibitem{nguyen2020variational}
Quoc~Phong Nguyen, Bryan Kian~Hsiang Low, and Patrick Jaillet.
\newblock Variational bayesian unlearning.
\newblock {\em arXiv preprint arXiv:2010.12883}, 2020.

\bibitem{agarwal2016second}
Naman Agarwal, Brian Bullins, and Elad Hazan.
\newblock Second-order stochastic optimization in linear time.
\newblock {\em stat}, 1050:15, 2016.

\bibitem{kingma2014adam}
Diederik~P Kingma and Jimmy Ba.
\newblock Adam: A method for stochastic optimization.
\newblock {\em arXiv preprint arXiv:1412.6980}, 2014.

\bibitem{northcutt2021confidentlearning}
Curtis~G. Northcutt, Lu~Jiang, and Isaac~L. Chuang.
\newblock Confident learning: Estimating uncertainty in dataset labels.
\newblock {\em Journal of Artificial Intelligence Research (JAIR)},
  70:1373--1411, 2021.

\bibitem{goodfellow2014explaining}
Ian~J Goodfellow, Jonathon Shlens, and Christian Szegedy.
\newblock Explaining and harnessing adversarial examples.
\newblock {\em arXiv preprint arXiv:1412.6572}, 2014.

\bibitem{he2016deep}
Kaiming He, Xiangyu Zhang, Shaoqing Ren, and Jian Sun.
\newblock Deep residual learning for image recognition.
\newblock In {\em Proceedings of the IEEE conference on computer vision and
  pattern recognition}, pages 770--778, 2016.

\end{thebibliography}

\newpage
\appendix
\title{Fixing Neural Networks by Leaving the Right Past Behind: \\ Supplementary Material}
\section{Algorithmic details}

\subsection{Problem formulation for model repairment}
\label{app:problem_formulation}

This section extends the discussion on the problem formulation of model repairment. Specifically, under the modelling assumptions presented in the main text, the goals for model repairment are the following:
\begin{itemize}
    \item For a set of ``failure cases'' $\mathcal{F} = \{ \bm{z}_f = (\bm{x}, y) \}$ where the model with (Bayesian) predictive inference makes wrong predictions, repair the model to make correct predictions on $\mathcal{F}$ and similar cases. 
    \item For a set of ``benchmark cases'' $\mathcal{B} = \{ \bm{z}_b = (\bm{x}, y)\}$, maintain a given level of prediction accuracy after model repairment.
\end{itemize}

There are further considerations for executing and evaluating model repairment in practice.
\paragraph{Generalisation and efficiency of model repairment}

In practice the number of failure cases might be large or even infinite. For example, an image classification model that fails on a test input with Gaussian noise may also fail on all the other inputs with such level of noise. So we need to consider both, \emph{generalisation} and \emph{efficiency} aspects of model repairment. Here, good generalisation means that the failure is fixed, not only for observed failure cases but also for future cases for which the model would make the same type of failure before fixes. On the other hand, efficiency of repairment considers the number of failure examples required for the model repairment method to fix a particular type of failure.

To describe both concepts in more details, in addition to a set of known/observed failure examples $\mathcal{F}$ collected by the users, we need to define the set of \emph{unknown/unobserved} failure examples $\mathcal{F}_U$. If examples in $\mathcal{F}$ and $\mathcal{F}_U$ are similar, then good \emph{generalisation} of a model repairment algorithm means that a model repaired by such method using information from $\mathcal{F}$ should produce correct predictions for instances in $\mathcal{F}_U$. On the other hand, a model repairment method is \emph{efficient} if it only needs a small set $\mathcal{F}$ of collected failure cases to achieve good generalisation of repairment.

\paragraph{Specificity of model repairment}
Furthermore, there may be multiple different scenarios in which the model fails, and therefore the set of failure cases may consist of several groups, i.e.,  
\begin{equation}
\mathcal{F} = \sqcup_{m=1}^M \mathcal{F}^{(m)}, \quad \mathcal{F}_U = \sqcup_{m=1}^M \mathcal{F}_U^{(m)},
\end{equation}
where $\mathcal{F}^{(m)}$ denotes the observed failure cases of failure type $m$ and $\mathcal{F}_U^{(m)}$ represents unobserved failure cases of the same type.
Targeting a specific type of mistake and repairing it one at a time may be desirable in practice. One type of mistake might incur more costs than others (e.g., in medical applications, false negative is generally more costly than false positive), so users might have different priorities for different types of errors to be fixed. This is especially the case when there exists a trade-off between fixing different types of errors, again the false negative vs false positive trade-off is a prevalent example.

\paragraph{Repairment by identifying and removing detrimental training data}
There can be many different reasons for a model with Bayesian predictive inference making wrong predictions on $\mathcal{F}$. In this work, we assume that \emph{the main reason is due to the existence of detrimental datapoints in $\mathcal{D}$}. Our hypothesis is that, by removing/correcting these detrimental datapoints and adapting the model with them, the model can be repaired to return correct labels for datapoints in $\mathcal{F}$. Base on the above hypothesis, the model repairment process contains the following steps:
\begin{enumerate}
    \item \textbf{Cause identification:} Identify a set of detrimental data points $\mathcal{C}$ in the training data $\mathcal{D}$ that contributed the most to the failure set $\mathcal{F}$.
    \item \textbf{Treatment: } Given the set of failure causes $\mathcal{C}$, adapt the model to predict correctly on the failure set $\mathcal{F}$, while maintaining performance on other examples which were correctly predicted previously. 
\end{enumerate}
% MEL: This whole previous paragraph is fully redundant with main text, consider deletion.

\subsection{The ``predictive approach'' for cause identification} \label{sec:supp_derivations}
We use the following function to describe the contribution of a subset $\mathcal{C} \subset \mathcal{D}$ to the model failures on examples in $\mathcal{F}$:
\begin{equation}
    r(\mathcal{C}) =\log \left(\frac{p(\mathcal{F} \given \mathcal{D} \setminus \mathcal{C})}{p(\mathcal{F} \given \mathcal{D})} \right).
\end{equation}
A naive approach would compute $p(\boldtheta \given \mathcal{D} \setminus \mathcal{C})$ for all subsets of $\mathcal{D}$ with all possible correction methods, which is prohibitively expensive. Instead we present a \textbf{``predictive'' approach} that removes this computational burden. First, notice that we make the i.i.d.~modelling assumption which means that one can write the likelihood term as  follows:
\begin{equation}
    p(\mathcal{D} \given \boldtheta) = p(\mathcal{D} \setminus \mathcal{C} \given \boldtheta) p(\mathcal{C} \given \boldtheta), \quad \forall \; \mathcal{C} \subset \mathcal{D}.
\label{eq:iid_factorisation}
\end{equation}
This allows us to expand the log evidence $\log \, p(\mathcal{F} \given \mathcal{D} \setminus \mathcal{C})$ as
\begin{equation*}
\begin{aligned}
\log p(\mathcal{F} \given \mathcal{D} \setminus \mathcal{C}) &= \log \int p(\mathcal{F} \given \boldtheta) p(\boldtheta \given \mathcal{D} \setminus \mathcal{C}) d\boldtheta \\
&= \log \int p(\mathcal{F} \given \boldtheta) \textcolor{red}{\frac{p(\mathcal{D} \setminus \mathcal{C} \given \boldtheta) p(\boldtheta)}{p(\mathcal{D} \setminus \mathcal{C})}} d\boldtheta \quad\quad \text{(Bayes' rule)} \\
&= \log \int p(\mathcal{F} \given \boldtheta) \frac{ \textcolor{red}{p(\mathcal{D} \given \boldtheta)} p(\boldtheta)}{ \textcolor{red}{p(\mathcal{C} \given \boldtheta) } p(\mathcal{D} \setminus \mathcal{C})} d\boldtheta \quad\quad \text{(by Eq.~(\ref{eq:iid_factorisation}))} \\
&= \log \int \frac{\textcolor{red}{p(\mathcal{D})}}{p(\mathcal{D}\setminus \mathcal{C})}\cdot \frac{p(\mathcal{F}\given \boldtheta) }{p(\mathcal{C}\given \boldtheta)} \cdot \frac{p( \mathcal{D}, \boldtheta)}{\textcolor{red}{p(\mathcal{D})}} d\boldtheta \quad\quad \text{(multiplying $\frac{p(\mathcal{D})}{p(\mathcal{D})}$ and rearranging terms)}\\
&= \log \int \frac{p(\mathcal{F} \given \boldtheta)}{p(\mathcal{C}\given \boldtheta)} \textcolor{red}{p(\boldtheta \given \mathcal{D})} d\boldtheta + \log \frac{p(\mathcal{D})}{p(\mathcal{D}\setminus \mathcal{C})}. \quad\quad \text{(Bayes' rule)}
\end{aligned}
\end{equation*}
Then we can rewrite the log density ratio as
\begin{equation}
\begin{aligned}
r(\mathcal{C})&=\log \left(\frac{p(\mathcal{F} \given \mathcal{D} \setminus \mathcal{C})}{p(\mathcal{F} \given \mathcal{D})} \right) 
\\ &= \log \int \frac{p(\mathcal{F} \given \boldtheta)}{p(\mathcal{C}\given \boldtheta)} p(\boldtheta \given \mathcal{D}) d\boldtheta 
- \log \, p(\mathcal{F} \given \mathcal{D}) 
+ \log p(\mathcal{D}) \textcolor{red}{- \log \int p(\mathcal{D} \setminus \mathcal{C} \given \boldtheta) p(\boldtheta) d\boldtheta} \\
& \qquad\qquad\qquad\qquad\qquad\qquad\qquad\qquad\qquad\qquad\qquad\qquad \text{(by definition of marginal distributions)}\\
\\ &= \log \int \frac{1}{p(\mathcal{C}\given \boldtheta)} \frac{p(\mathcal{F} \given \boldtheta) p(\boldtheta \given \mathcal{D})}{p(\mathcal{F} \given \mathcal{D})}  d\boldtheta 
- \log \int \frac{\textcolor{red}{p(\mathcal{D} \given \boldtheta)}}{\textcolor{red}{p(\mathcal{C} \given \boldtheta)}} \frac{p(\boldtheta)}{p(\mathcal{D})} d\boldtheta \\
& \qquad\qquad\qquad\qquad\qquad\qquad\qquad\qquad\qquad\qquad\qquad\qquad \text{(by Eq.~(\ref{eq:iid_factorisation}) and rearranging terms)}\\
&= \log \int \frac{ \textcolor{red}{p(\boldtheta \given \mathcal{D}, \mathcal{F})}}{p(\mathcal{C}\given \boldtheta)} d\boldtheta - \log \int \frac{\textcolor{red}{p(\boldtheta \given \mathcal{D})}}{p(\mathcal{C}\given \boldtheta)}  d\boldtheta . \quad\quad \text{(Bayes' rule)}
\end{aligned}
\end{equation}
By doing so, instead of computing $p(\boldtheta \given \mathcal{D} \setminus \mathcal{C})$ for every possible subset $\mathcal{C}$, the ``predictive approach'' only requires computing $p(\boldtheta \given \mathcal{D}, \mathcal{F})$ once. As shown in the main text, with (approximations of) the two posteriors $p(\boldtheta \given \mathcal{D})$ and $p(\boldtheta \given \mathcal{D}, \mathcal{F})$ at hand, the log density ratio $r(\mathcal{C})$ can be efficiently approximated by Monte Carlo and/or further approximations described in the main text that employ Taylor expansions. This approach is ``predictive'' in the sense that the influence of candidate set $\mathcal{C}$ is evaluated by computing ``predictions'' $p(\mathcal{C} \given \boldtheta)^{-1}$ on them using the two posterior distributions, which is different from existing approaches that compute predictions on $\mathcal{F}$ using approximations to the modified posterior $p(\boldtheta \given \mathcal{D} \setminus \mathcal{C})$.

\subsection{Objective for EWC-influence}
\label{app:ewc_identification_objective}
Recall in the main text the first-order Taylor series approximation to $r(\mathcal{C})$ is
\begin{equation*}
    \begin{aligned}
    r(\mathcal{C}) \approx \sum_{\bz \in \mathcal{C}} \ \hat{r}(\bz), \quad \hat{r}(\bz)= \mathbb{E}_{p(\boldtheta \given \mathcal{D})}\left[\log p(\bz \given \boldtheta) \right] - \mathbb{E}_{p(\boldtheta \given \mathcal{D}, \mathcal{F})}\left[\log p(\bz \given \boldtheta) \right]. 
    \end{aligned}    
\end{equation*}
Therefore the ``predictive approach'' for computing $r(\mathcal{C})$ as well as the approximated form require the computation of (approximate) posteriors $p(\boldtheta \given \mathcal{D})$ and $p(\boldtheta \given \mathcal{D}, \mathcal{F})$. This can be achieved using continual learning: we assume the model has been trained on $\mathcal{D}$ and an approximation $q(\boldtheta) \approx p(\boldtheta \given \mathcal{D})$ has been obtained. Then the current task for continual learning is to adapt the trained model on the failure cases $\mathcal{F}$, which leads to an adapted approximation $q\DF(\boldtheta) \approx p(\boldtheta \given \mathcal{D}, \mathcal{F})$.

Elastic Weight Consolidation (EWC) \citep{kirkpatrick2017overcoming} is a continual learning algorithm that can be interpreted as updating the maximum a posteriori (MAP) approximation to the posterior given new tasks. To see this, first in this approach the $q$ posteriors are assumed to be delta measures. In other words, $q(\boldtheta) = \delta(\boldtheta - \hat{\boldtheta})$ where $\hat{\boldtheta}$ is the parameters of the trained model, and $q\DF(\boldtheta) = \delta(\boldtheta - \hat{\boldtheta}\DF)$ where as we shall see $\hat{\boldtheta}\DF$ is the parameter obtained by running EWC using $\mathcal{F}$. With these assumptions, the EWC-influence is defined as:
\begin{equation*}
    \begin{aligned}
    r(\mathcal{C}) \approx \sum_{\bz \in \mathcal{C}} \ \hat{r}(\bz) \approx \sum_{\bz \in \mathcal{C}} \ \tilde{r}(\bz), \quad \tilde{r}(\bz)= \log p(\bz \given \hat{\boldtheta})  - \log p(\bz \given \hat{\boldtheta}\DF).
    \end{aligned}    
\end{equation*}

It remains to discuss the optimisation procedure for obtaining $\hat{\boldtheta}\DF$. As motivated, we consider MAP approximations to the posterior, which seeks to find the maximum of the log posterior $\log \, p(\boldtheta \given \mathcal{D}, \mathcal{F})$. Notice that by Bayes' rule:
\begin{equation*}
    p(\boldtheta \given \mathcal{D}, \mathcal{F}) \propto p(\mathcal{F} \given \boldtheta) p(\boldtheta \given \mathcal{D}) \quad \Rightarrow \quad \log \, p(\boldtheta \given \mathcal{D}, \mathcal{F}) = \log \, p(\mathcal{F} \given \boldtheta ) + \log \, p(\boldtheta \given \mathcal{D}) + \text{constant}.
\end{equation*}
Computing the first term $\log \, p(\mathcal{F} \given \boldtheta)$ in the MAP objective is straightforward given the i.i.d.~modelling assumption. For the second term, as $\log \, p(\boldtheta \given \mathcal{D})$ is intractable, the EWC approach constructs a Laplace approximation to it by assuming the trained model parameter $\hat{\boldtheta}$ as a MAP point estimate of $p(\boldtheta \given \mathcal{D})$. In detail, a Laplace approximation to the posterior is
\begin{equation} \label{eq:taylor_expansion}
\log \, p(\boldtheta\given \mathcal{D}) \approx \frac{1}{2}(\boldtheta - \hat{\boldtheta})^\top \boldH[\log \, p(\hat{\boldtheta} \given \mathcal{D})] (\boldtheta - \hat{\boldtheta}) + \text{constant}.
\end{equation}
where $\boldH[f(\boldtheta)]$ denotes the Hessian matrix of a twice-differentiable function $f(\boldtheta)$ with respect to parameters $\boldtheta$. Further decomposing the Hessian term yields: 
\begin{align}
    \boldH[\log \, p(\hat{\boldtheta} \given \mathcal{D})] & =  \boldH[\log \, p(\mathcal{D}\given \hat{\boldtheta} ) + \log \, p(\hat{\boldtheta}) - \log \, p(\mathcal{D})]\\
    & = \underbrace{\boldH[\log p(\mathcal{D}\given \hat{\boldtheta} )]}_{\approx -N\hat{\boldF}_{\hat{\boldtheta}}} + \underbrace{\boldH[\log p(\hat{\boldtheta})]}_{\text{Prior term}}
\end{align}
where $N$ is the total number of samples in $\mathcal{D}$ and $\hat{\boldF}_{\hat{\boldtheta}}$ is an empirical estimate of the \emph{Fisher information matrix}. Assuming the zero-mean isotropic Gaussian prior $p(\boldtheta)$ with precision $\lambda\boldsymbol{I}$,  the second term becomes $\boldH[\log p(\hat{\boldtheta})] = -\lambda \boldsymbol{I}$. Substituting these results back into Eq.~(\ref{eq:taylor_expansion}) gives us: 
\begin{equation}
    \log \, p(\boldtheta\given \mathcal{D}) \approx - \frac{N}{2}(\boldtheta - \hat{\boldtheta})^\top \hat{\boldF}_{\hat{\boldtheta}}  (\boldtheta - \hat{\boldtheta}) - \frac{\lambda}{2}||\boldtheta - \hat{\boldtheta}||^2_2 + \text{constant}.
\end{equation}
Combining terms, the optimisation task for EWC adaptation using failure set is:
\begin{equation}
    \hat{\boldtheta}\DF = \arg\max_{\boldtheta} \ \log \, p(\mathcal{F} \given \boldtheta) - \frac{N}{2}(\boldtheta - \hat{\boldtheta})^\top \hat{\boldF}_{\hat{\boldtheta}}  (\boldtheta - \hat{\boldtheta}) - \frac{\lambda}{2}||\boldtheta - \hat{\boldtheta}||^2_2,
\end{equation}
which is as presented in the main text.
We further note that in the original EWC formulation, the empirical Fisher information is further simplified to contain \textit{diagonal} entries only --- with millions of model parameters as is typically the case with neural networks, as computing the full matrix is expensive. 

\subsection{Objective for EWC-deletion}
\label{app:ewc_deletion_objective}
Assuming that the detrimental datapoints $\mathcal{C} \subset \mathcal{D}$ have been identified, the next step for model repairment is \emph{treatment} where we seek to remove the influence of $\mathcal{C}$ to the model. This is done by computing (an approximation to) $p(\boldtheta \given \mathcal{D} \setminus \mathcal{C})$ which can also be done via continue learning. To see this, we can show by using Bayes' rule and Eq.~(\ref{eq:iid_factorisation}) again,
\begin{equation}
    \log p(\boldtheta \given \mathcal{D} \setminus \mathcal{C}) = \log \frac{p(\mathcal{D} \setminus \mathcal{C} \given \boldtheta)p(\boldtheta)}{p(\mathcal{D} \setminus \mathcal{C})} = \log \frac{1}{p(\mathcal{C}\given \boldtheta)} + \log \underbrace{\frac{p(\mathcal{D}\given \boldtheta)p(\boldtheta)}{p(\mathcal{D})}}_{= p(\boldtheta \given \mathcal{D})} + \log \frac{p(\mathcal{D})}{p(\mathcal{D} \setminus \mathcal{C})}.
\end{equation}
This means that finding the MAP estimate of $p(\boldtheta \given \mathcal{D} \setminus \mathcal{C})$ is equivalent to maximising $-\log p(\mathcal{C} \given \boldtheta) + \log p(\boldtheta \given \mathcal{D})$ w.r.t.~$\boldtheta$. With further approximation to the posterior $p(\boldtheta \given \mathcal{D})$ using the Laplace method as discussed in section \ref{app:ewc_identification_objective}, one can write the optimisation task for EWC-deletion as presented in the main text, namely:
\begin{equation}
    \hat{\boldtheta}\DmC = \arg\max_{\boldtheta} \ -\log p(\mathcal{C} \given \boldtheta) - \frac{N}{2}(\boldtheta - \hat{\boldtheta})^\top \hat{\boldF}_{\hat{\boldtheta}}  (\boldtheta - \hat{\boldtheta}) - \frac{\lambda}{2}||\boldtheta - \hat{\boldtheta}||^2_2.
\end{equation}

%%%%%%%%%%%%%%%%%%%%%%%%%%%%%
\section{Additional Results} \label{sec:supp_extra_results} 

\subsection{Sample Efficiency of Cause Identification}
\begin{figure}[h!]
\centering
  \includegraphics[width=0.95\linewidth]{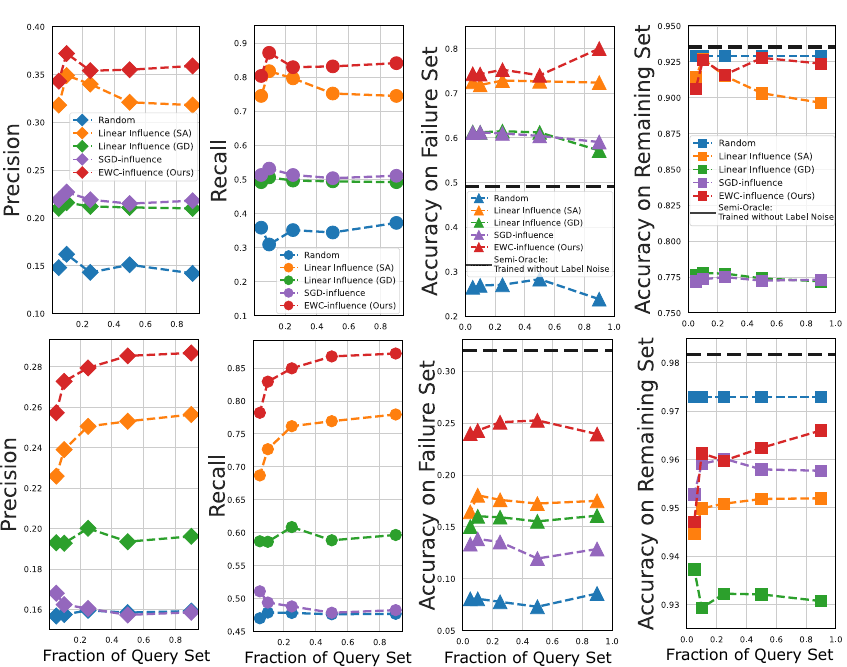}
%   \vspace{-2mm}
  \caption{\footnotesize %{\bf EWC-influence remains competitive as we decrease the size of the failure set.}
  Comparison of sample efficiency for cause identification performance on (top) MNIST and (bottom) CIFAR-10 datasets with class-dependent annotation noise. On the x-axis of each sub-figure, we vary the size of the query failure set $\mathcal{F}_q$ used for cause identification. From left to right, (a) precision, (b) recall, (c) accuracy on holdout failure set $\mathcal{F}_h$, and (d) accuracy on the remaining test set. %\MFP{(Add amount of causes removed, which operation point in Fig. 2)}.
  }
  \label{fig:sample_efficiency}
\end{figure}

Fig.~\ref{fig:sample_efficiency} compares the sample efficiency of different approaches for cause identification on MNIST and CIFAR-10 datasets with annotation noise. More specifically, we want to understand how many failure cases in the query set $\mathcal{F}_q$ we need to see for cause identification to work, i.e., how performance of each approach varies as we reduce %\MFP{It might be more intuitive if we were to plot mirrored x-axis, such that left side corresponds to original query set (assuming that is the size used for Figure 2.)}
the size of the query set. Here, all detrimental examples i.e., $\tilde{r}(\bz)<0$ are removed from the training data and the corresponding metrics are measured.  EWC-influence approach performs the best in terms of precision, recall, and accuracy on holdout failure set $\mathcal{F}_h$ even as we decrease the size of the query set. All approaches exhibit a stable behavior w.r.t the size of query set. The biggest drop occurs in accuracy on the remaining test set $\mathcal{T}\setminus\mathcal{F}$: EWC-influence presents a stronger degradation in small sample regimes (around 5\% of the original failure set) but without dropping below other approaches.

For the failure accuracy on MNIST, even when the query set is as small as 5\% of the original one ($\approx 20$ examples), all approaches display improvement over the semi-oracle that is trained after removing all instances of annotation errors: this means that all methods, in a sample-efficient manner, not only correct the synthetically added annotation errors, but they also remove other harmful examples that are naturally present to begin with.

% For MNIST, all approaches improves above the semi-oracle\footnote{Note that the semi-oracle considered here is different from the one considered in Fig.~\ref{fig:labelnose_detection} and Fig.~\ref{fig:input_noise_detection}: the semi-oracle here corresponds to training with the uncorrupted original dataset, whereas oracle in the context of the other figures refer to training with corrupted datasets and assuming that we already know which datapoints are the corrupted ones.} that trains with an uncorrupted training set: this means that all methods not only correct the synthetically added examples, but they also remove other harmful examples that are naturally present to begin with.

% Why do you call it semi-oracle and not oracle? Potential point of confusion is that Oracle in this Figure is different from the concept of Oracle in Fig. 2, so added footnote.

\clearpage
\subsection{Speed Comparison}
Table~\ref{tab:computation} shows the total computation time of the proposed EWC-influence and other baselines for cause identification on a single Tesla K80 GPU with 12GB of RAM. Note that we rely on the publicly available implementation of SGD-influence\footnote{\url{https://github.com/sato9hara/sgd-influence}}, and that we have implemented our own version of linear influence functions in Pytorch.
Overall, EWC-Influence is either as fast or faster as other baselines, achieving one order of magnitude speed boost compared to SGD-influence in all cases. For CIFAR10 where a considerably larger base model is used, EWC-influence is consistently faster than other approaches, around twice faster than linear influence methods. For MNIST, EWC-influence attains similar computation time as linear influence approaches. 

\begin{table}[ht!]
	\caption {Comparison of total computation time for cause identification.}\label{tab:computation}
 	\label{tab:speed_comparison}
    \center
	\begin{tabular}{|l|c|c|c|}
		\hline
		\multicolumn{1}{|c}{\textbf{Experiment}} &  \multicolumn{1}{|c|}{\textbf{Method}} & \multicolumn{1}{c|}{\textbf{Time (s)}} & \multicolumn{1}{c|}{\textbf{Time (s)}} \\
		&     & ( MNIST ) & (CIFAR10)  \\
		\hline
	            & Linear Influence (SA) & 16.4 & 1322.4\\
        Label Noise & Linear Influence (GD) & 11.2 & 996.2 \\
	              & SGD-Influence         & 185.1 & 8301.1 \\
	                      & EWC-Influence (Ours)  & \textbf{9.8 }& \textbf{496.3 }\\
	   \hline
                              & Linear Influence (SA) & 10.8   & 1141.1  \\
              Random          & Linear Influence (GD) &  15.7  & 978.4 \\
	        Input Noise  & SGD-Influence              & 188.2    & 8285.6 \\
	                   & EWC-Influence (\textbf{Ours})& 11.0    & \textbf{527.5}\\
	    \hline
                           &  Linear Influence (SA) & 14.3   & 1312.0  \\
        Adversarial       & Linear Influence (GD)   & \textbf{10.0  } & 984.2 \\
            Poisoning  & SGD-Influence              &  177.29  & 7803.7\\
                              & EWC-Influence (\textbf{Ours})  & 14.5 & \textbf{528.9} \\
        \hline
	\end{tabular}
	\vspace{-2mm}
% 	\vspace{-2mm}
\end{table}

\subsection{Qualitative Results}

\begin{figure}[ht!]
\centering
  \includegraphics[width=0.95\linewidth]{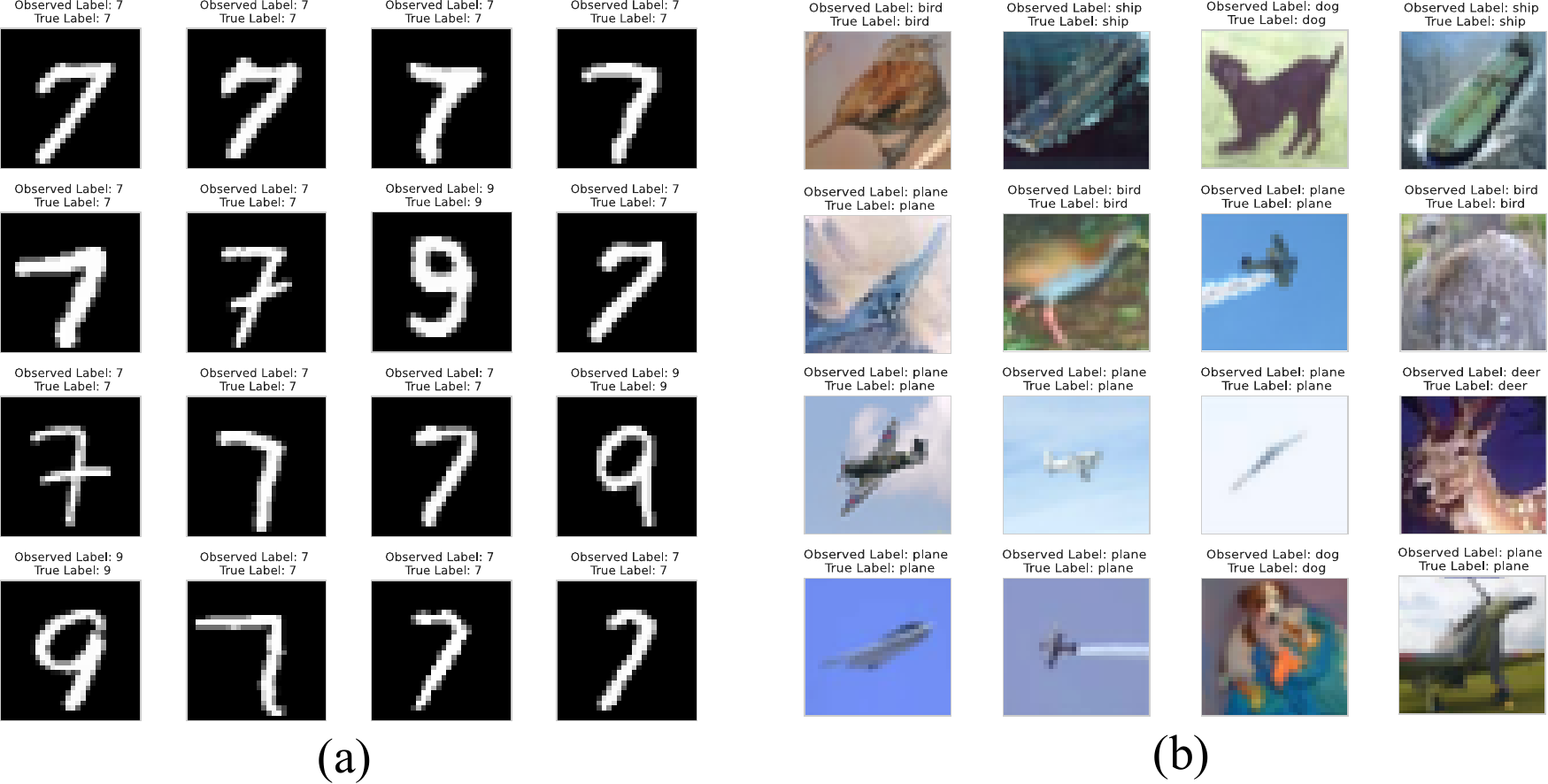}
  \caption{\footnotesize Examples of 16 \textbf{least} harmful examples for MNIST and CIFAR-10 with annotation noise as ranked by EWC-influence. None of the selected examples were corrupted with annotation noise.}
\label{fig:label_noise_least_harmful}
\end{figure}

\paragraph{Annotation noise.} In the main text, Fig.~\ref{fig:labelnose_detection}(d) shows examples with annotation noise ranked as most harmful according to EWC-influence. Similarly, Fig.~\ref{fig:label_noise_least_harmful} shows examples ranked as least harmful according to EWC-influence. All of these examples correspond to non-corrupted examples with correct labels. 

\begin{figure}[h!]
\centering
  \includegraphics[width=0.95\linewidth]{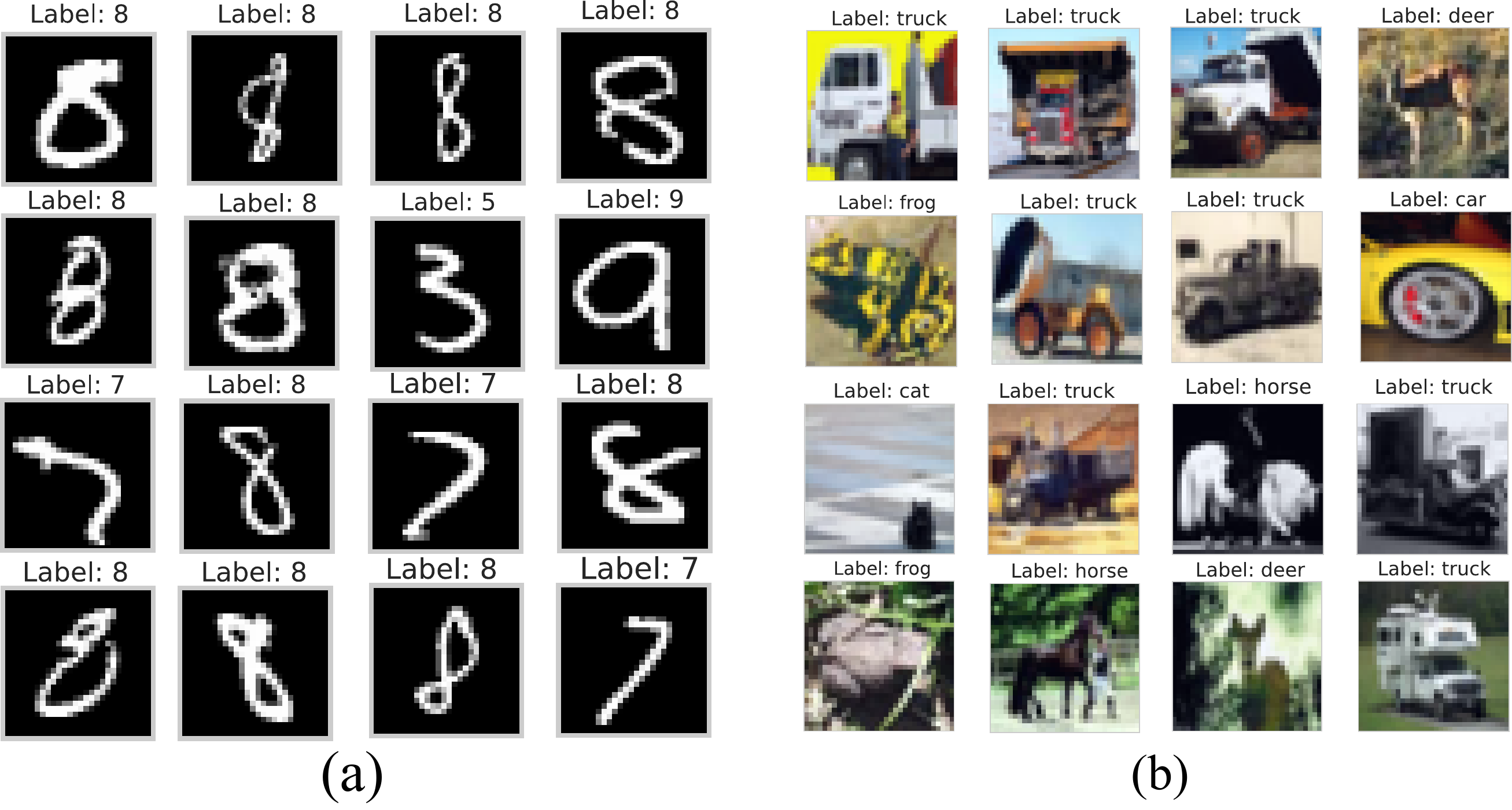}
%   \vspace{-4mm}
  \caption{\footnotesize Examples of 16 most harmful examples for MNIST and CIFAR-10 with random input noise as ranked by EWC-influence. None of the selected examples were corrupted with random input noise. 
  }
  \label{fig:input_not_harmful}
\end{figure}

\clearpage
\paragraph{Random input noise.} Fig.~\ref{fig:input_not_harmful} shows examples that are ranked highest (most harmful) by EWC-influence for datasets contaminated with random input noise. Recall that the target classes of the input noise are 1, 6, 7, 9 for MNIST and plane, bird, cat, dog for CIFAR10, and the rest of the images are free of such noise. First of all, we observe that the most detrimental examples belong to the non-target classes. As shown in the main text, while the input noise itself may not harm the performance of the model by much, the sample size of clean images in the target classes is still smaller as a result of noise injection---such group imbalance can be rectified by sub-sampling the dominant group, for example, by removing those identified detrimental data points. Secondly, many of them appear to be ambiguous instances in non-target classes. It is worth noting that for MNIST, one of the identified examples is interestingly a real instance of 3 that is incorrectly labelled as 5, which is also reported by Northcutt \etal \cite{northcutt2021confidentlearning}.

%As discussed in the main text, EWC-influence does not pick up examples that are corrupted with input noise  because; in contrast, EWC-influence rank as harmful uncorrupted images that do present unconventional shapes that could easily be miss-classified as other classes.

%%%%%%%%%%%%%%%%%% (RYU) Include this result if reviewers ask for it, perhaps or in the camera-ready. I'm worried that it can cause confusion. 
% %%% 
\subsection{Treating Models by Forgetting Detrimental Past and Learning from Present Mistakes}
\begin{figure}[ht!]
\centering
  \vspace{-3mm}
  \includegraphics[width=0.75\linewidth]{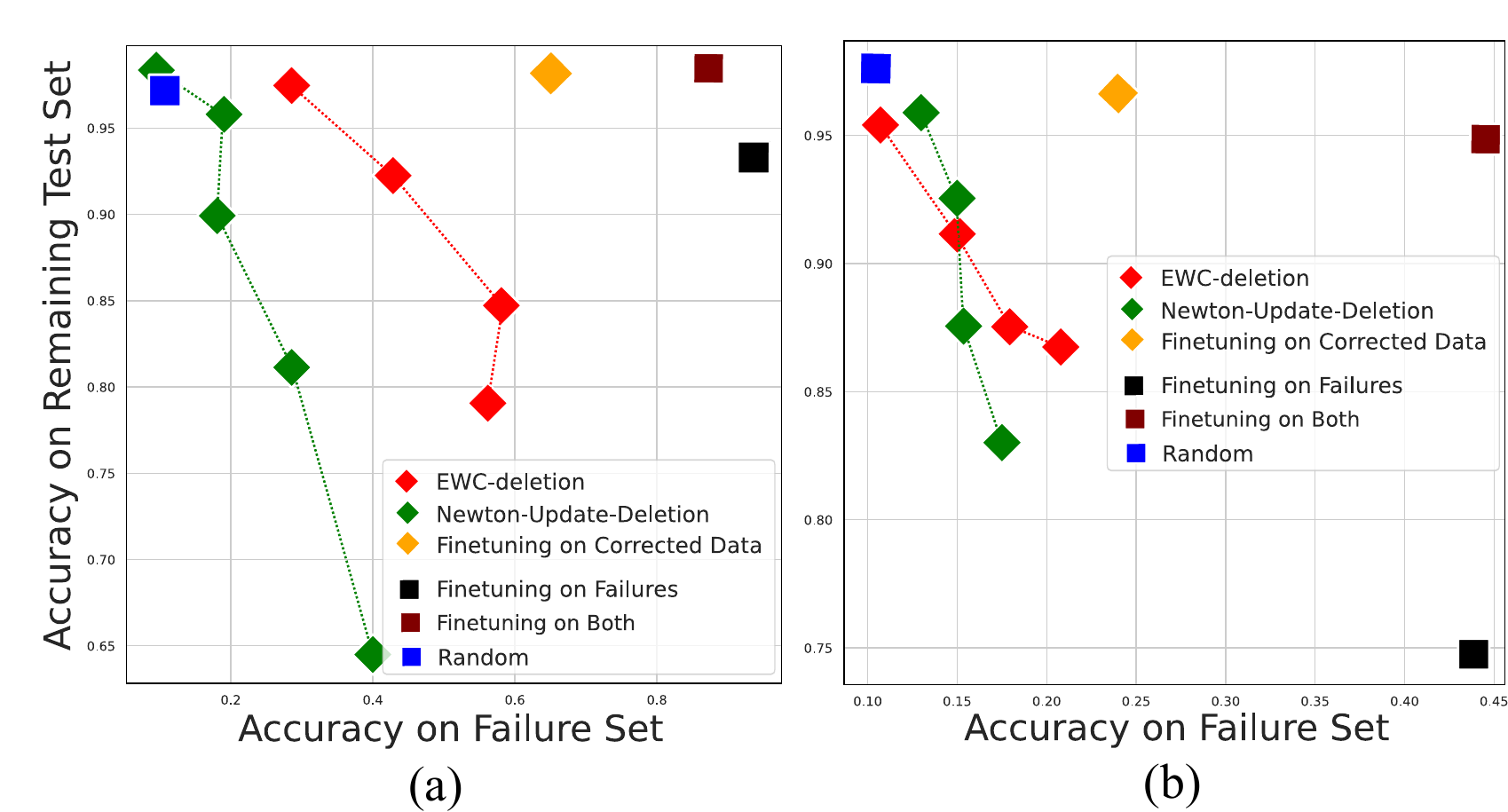}
  \vspace{-3mm}
  \caption{\footnotesize Comparison of deletion-based treatment methods and the direct fine-tuning on the failures on (a) MNIST and (b) CIFAR-10 datasets with noisy annotations.}
  \label{fig:treatment_comparison_with_finetuning}
\end{figure}

While the primary goal of this work is to investigate how many prediction errors can be remedied by identifying harmful training data and removing them, one could alternatively use the labelled failure cases directly to adapt the model. Fig.~\ref{fig:treatment_comparison_with_finetuning} shows our preliminary results where we compare the deletion based methods to an approach that fine-tunes the model directly on the failure query set $\mathcal{F}_{q}$ with an L2-norm based locality constraint $||\boldtheta - \hat{\boldtheta}||^2_2$ \cite{zhu2020modifyingmemories} and its combination with the best deletion-based approach, that is, fine-tuning on the corrected dataset $\mathcal{D}\setminus \mathcal{C}$. As with other experiments, early stopping is performed based on the loss on a portion of the query set $\mathcal{F}_{q}$. We see that while fine-tuning on $\mathcal{F}_{q}$ (black points) leads to a higher accuracy (on the holdout failure set $\mathcal{F}_{h}$) than fine-tuning on $\mathcal{D}\setminus \mathcal{C}$ (yellow points), the accuracy on the remaining test set $\mathcal{T}\setminus \mathcal{F}$ is worse, even with the weight constraint~---~by 6\% on MNIST and 21\% on CIFAR-10. This issue of over-fitting to the failure cases is also reported in recent works such as \cite{cao2021editing,sinitsin2020editable}. Importantly, by combining the two approaches (brown points), we can attain the best trade-off between the failure and the maintenance accuracy, indicating the complementarity between the proposed data correction methods and such fine-tuning approaches.

\newpage
%%%%%%%%%%%%%%%%%%%%%%%%%
\section{Experimental Details} \label{sec:supp_experimental_details}

\paragraph{Datasets}
 we perform our experiments on the MNIST digit classification task and the CIFAR-10 object recognition task.  The MNIST dataset consists of $60,000$ training and $10,000$ testing examples, all of which are $28\times28$ grayscale images of digits from $0$ to $9$ ($10$ classes). The CIFAR-10 dataset consists of $50,000$ training and $10,000$ testing examples, all of which are $32\times32$ coloured natural images drawn from $10$ classes. Both datasets are preprocessed by subtracting the mean, but no data augmentation is used. For MNIST, to make the task more challenging, we randomly select 3000 examples from the training split and train the base models while the entire test set is used for evaluation.

\paragraph{Architecture Details} 
For MNIST, the base classifier was defined as a CNN architecture comprised of $4$ convolution layers, each with $3\times3$ kernels follower by Relu. The number of kernels in respective layers are $\{32, 32, 64, 64\}$. After the first two convolution layers, we perform $2\times2$ max-pooling, and after the last one, we further down-sample the features with Global Average Pooling (GAP) prior to the final fully connected layer. For CIFAR-10, we used a 50-layer ResNet \cite{he2016deep}. 

\paragraph{Optimisation}
For all experiments, we employ the same training scheme unless otherwise stated. We optimize parameters using Adam \cite{kingma2014adam} with initial learning rate of $10^{-3}$ and $\beta = [0.9, 0.999]$, with minibatches of size $64$ and train for max $100$ epochs with early stopping with a patience of 5, that is, training is stopped after 5 epochs of
no progress on the validation set (10\% of the training set). For computing both EWC-influence and EWC-deletion, we also employed the same training scheme but applied early stopping based on the performance on a validation split (10\%) of the failure query set $\mathcal{F}_{q}$. 

In Fig.~\ref{fig:deletion_comparison} in Sec.~\ref{sec:results_treatment}, we present the performance of Newton update removal and EWC-deletion (ours) with different hyper-parameter settings. For Newton-update deletion \cite{guo2019certified}, we vary the step size $\gamma>0$ of the gradient ascent by scaling the second term in Eq.~\eqref{eq:koh_data_deletion}. For EWC-deletion (our method), we vary the amount of weight regularisation --- the second term in Eq.~\eqref{eq:deletion_ewc} --- by scaling it by $2/\gamma N$ where $N$ denotes the number of training datapoints, and $\gamma>0$. We also set the strength of the prior term to $\lambda=0$. We run both methods for different values of $\gamma$ in the range $[0.01, 0.05]$.

\newpage
\section{Enlarged Figures}\label{sec:supp_enlarged_figs}

\begin{sidewaysfigure*}[ht!]
\centering
  \includegraphics[width=1.0\linewidth]{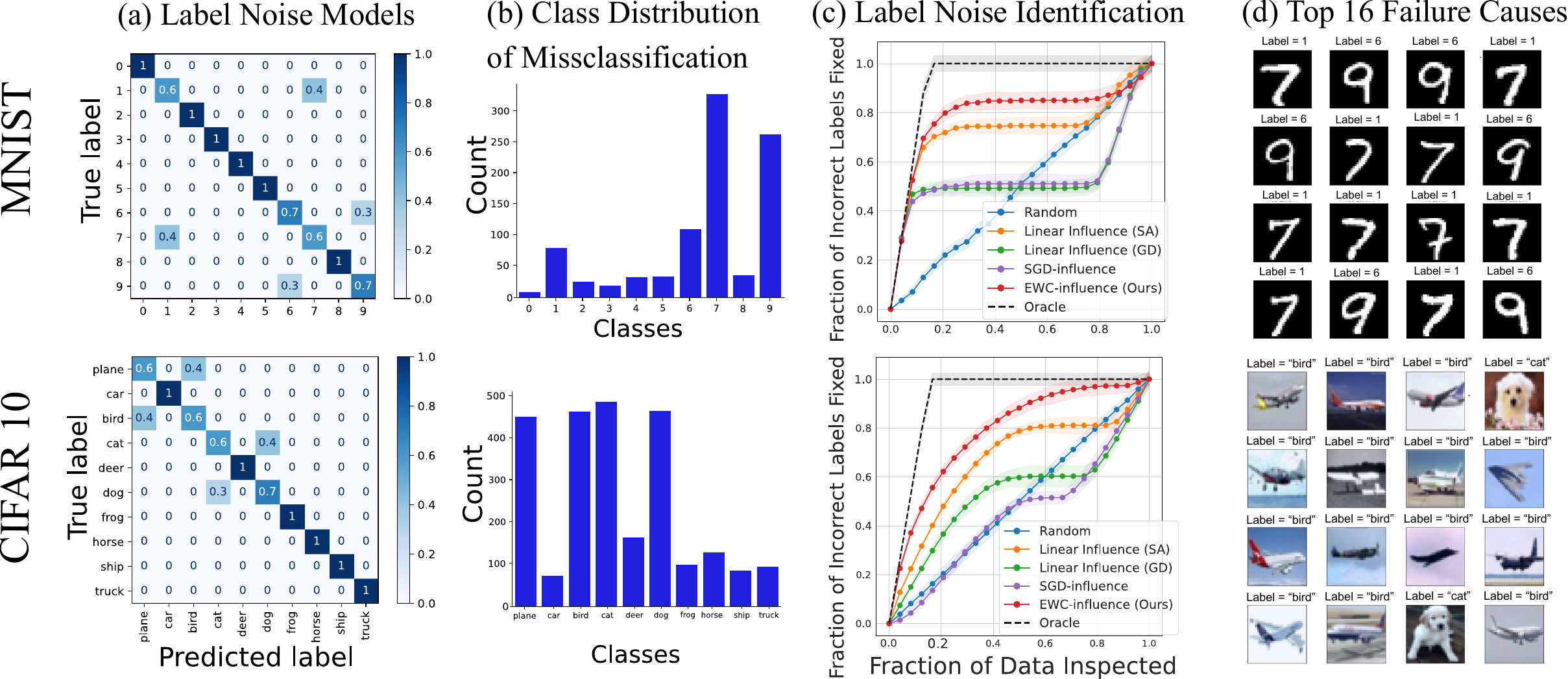}
  \caption{\footnotesize %{\bf EWC-influence outperforms other baselines at cause identification in the presence of annotation noise.}
  Results on cause identification in the presence of annotation noise.
  (a) shows the confusion matrices used to simulate class-dependent label noise on MNIST and CIFAR-10. (b) shows the class distribution of the misclassified examples for a single run. (c) plots how much of the  identified causes match the samples with incorrect labels for different approaches. The shade represents the standard deviation computed from 5 different runs. (d) shows the top 16 causes of the failures as ranked by EWC-influence. }
\end{sidewaysfigure*}

\begin{sidewaysfigure}[ht!]
\centering
  \includegraphics[width=1.0\linewidth]{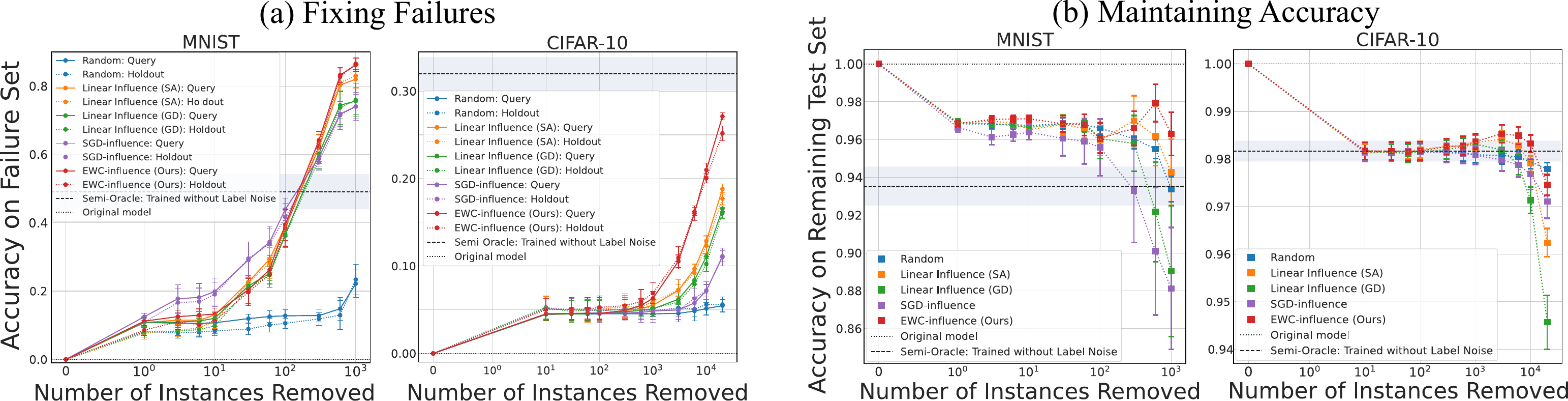}
  \caption{\footnotesize Comparison of the quality of identified causes in the presence of annotation noise. The impact of gradually removing samples in MNIST and CIFAR-10 datasets in the order of influence values $r(\bz)$ are measured on the failure sets (holdout and query) in (a), and on the remaining test set in (b). \blue{We note that in (b), accuracy values start at 1.0 as they are calculated on the set of test samples on which the original model makes correct predictions}. We also plot the performance of another reference (``semi-oracle'') that is the original model fine-tuned on the training data without the label noise instances. The means and standard deviations of all quantities are calculated over 5 different runs. }
\end{sidewaysfigure}

\begin{sidewaysfigure*}[t]
\centering
  \includegraphics[width=1.0\linewidth]{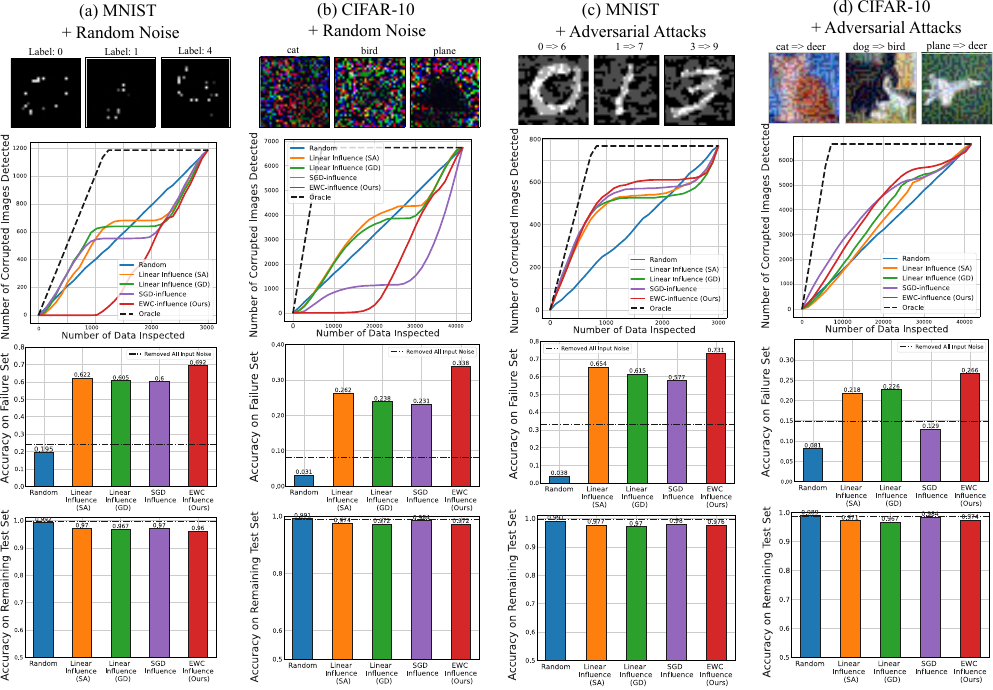}
  \caption{\footnotesize 
  Results on cause identification in the presence of different input noises. From top to bottom, we show i) examples of corrupted samples (synthetic proxy for potential causes of failure), ii) how many of the identified causes correspond to samples corrupted with input noise, iii) and iv) performance in failure holdout set $\mathcal{F}_{h}$ and remaining test set when removing the top 1000/20000 identified causes in MNIST/CIFAR-10. The influence values are calculated with respect to 50\% of test-time failure cases that belong to the classes that suffer from input noise. EWC-Influence identifies ``harmful'' (adversarial) input noise better than random while avoiding ``harmless'' (random) input noise.
  %\MFP{(Would it make sense to highlight in 2nd row the cuting plane, i.e., how many removed data does 3rd and 4th row correspond to?)}  
  %(a,b)  Random input noise for MNIST (top row) and CIFAR10 (bottom row) datasets. 
  }
 \end{sidewaysfigure*}

\end{document}